%% file: main.tex
\documentclass[10pt,twocolumn,letterpaper]{article}

\usepackage[pagenumbers]{cvpr} 


\definecolor{cvprblue}{rgb}{0.21,0.49,0.74}
\usepackage[pagebackref,breaklinks,colorlinks,allcolors=cvprblue]{hyperref}

\usepackage{algorithm}
\usepackage{algpseudocode}
\usepackage{multirow}

\usepackage{caption}
\usepackage{subcaption}

\algnewcommand{\LineComment}[1]{\State \(\triangleright\) #1}

\def\paperID{*****} 
\def\confName{CVPR}
\def\confYear{2025}

\title{Accelerating Multimodal Large Language Models by Searching Optimal \\ Vision Token Reduction}

\author{
Shiyu Zhao$^{1}$ $\quad$ Zhenting Wang$^{1}$ $\quad$ Felix Juefei-Xu$^{2}$ $\quad$ Xide Xia$^{2}$ $\quad$ Miao Liu$^{2}$ \\ $\quad$ Xiaofang Wang$^{2}$ $\quad$ Mingfu Liang$^{2}$ $\quad$ Ning Zhang$^{2}$ $\quad$ Dimitris N. Metaxas$^{1}$ $\quad$ Licheng Yu$^{2}$\\
\\
$^{1}$Rutgers University $\quad$ $^{2}$Meta
}

\begin{document}
\maketitle
\input{sec/0_abstract}

\input{sec/1_intro}
\input{sec/2_related_work}

\input{sec/3_method}

\input{sec/4_experiments}

\input{sec/5_conclusion}

{
    \small
    \bibliographystyle{ieeenat_fullname}
    \bibliography{main}
}

\input{sec/X_suppl}

\end{document}

%% file: sec/0_abstract.tex
\begin{abstract}
Prevailing Multimodal Large Language Models (MLLMs) encode the input image(s) as vision tokens and feed them into the language backbone, similar to how Large Language Models (LLMs) process the text tokens. 
However, the number of vision tokens increases quadratically as the image resolutions, leading to huge computational costs.
In this paper, we consider improving MLLM's efficiency from two scenarios, 
(\uppercase\expandafter{\romannumeral1}) Reducing computational cost without degrading the performance. 
(\uppercase\expandafter{\romannumeral2}) Improving the performance with given budgets.
We start with our main finding that the ranking of each vision token sorted by attention scores is similar in each layer except the first layer.
Based on it, we assume that the number of essential top vision tokens does not increase along layers.
Accordingly, for Scenario \uppercase\expandafter{\romannumeral1}, we propose a greedy search algorithm (G-Search) to find the least number of vision tokens to keep at each layer from the shallow to the deep.
Interestingly, G-Search is able to reach the optimal reduction strategy based on our assumption.
For Scenario \uppercase\expandafter{\romannumeral2}, based on the reduction strategy from G-Search, we design a parametric sigmoid function (P-Sigmoid) to guide the reduction at each layer of the MLLM, whose parameters are optimized by Bayesian Optimization.
Extensive experiments demonstrate that our approach can significantly accelerate those popular MLLMs, e.g. LLaVA, and InternVL2 models, by more than $2 \times$ without performance drops. 
Our approach also far outperforms other token reduction methods when budgets are limited, achieving a better trade-off between efficiency and effectiveness.

\end{abstract}

%% file: sec/1_intro.tex
\section{Introduction}
\label{sec:intro}

Multimodal Large Language Models (MLLMs) usually leverage a pre-trained vision encoder to encode the input image(s) into vision tokens that are fed into pre-trained Large Language Models (LLMs).
Recent studies~\citep{li2024monkey,xu2024llava} demonstrate that increasing the number of vision tokens by using high resolution inputs significantly enhances the effectiveness of MLLMs.
However, such large number of vision tokens leads to inefficiency of model inference, preventing MLLMs from real-world applications.

To address such inefficiency, prompt-agnostic approaches are proposed to reduce the number of vision tokens before feeding them into LLMs. 
Various approaches have been proposed, including local attention pooling~\citep{luo2024feast,li2024tokenpacker,li2024mini}, re-samplers as compression layers~\citep{xu2024llava,zhang2024beyond}, or deformable convolutions~\citep{cha2024honeybee}, etc.
One noticeable issue of prompt-agnostic approaches is the ignorance of the input text prompts from the user. 
Since different prompts may focus on different regions of the image, prompt-agnostic approaches are likely to preserve irrelevant vision tokens that can be potentially further removed.
As a remedy, prompt-aware methods are proposed to leverage text prompts in the vision token reduction. 
For example, FastV~\citep{chen2024image}, VTW~\citep{lin2024boosting}, and PDrop~\cite{xing2024pyramiddrop} remove vision tokens at certain layers of the LLM within the MLLM. 
However, their token reductions are all designed with handcrafted rules that vary by MLLMs. 
They also focus more on reducing the computational cost without performance drops, while not addressing how to assign computations for better effectiveness with limited budgets.
We believe the latter is in greater demand in practical edge-device applications.

\begin{figure*}[t]\centering
    \begin{subfigure}[b]{0.32\textwidth}
        \includegraphics[width=\textwidth]{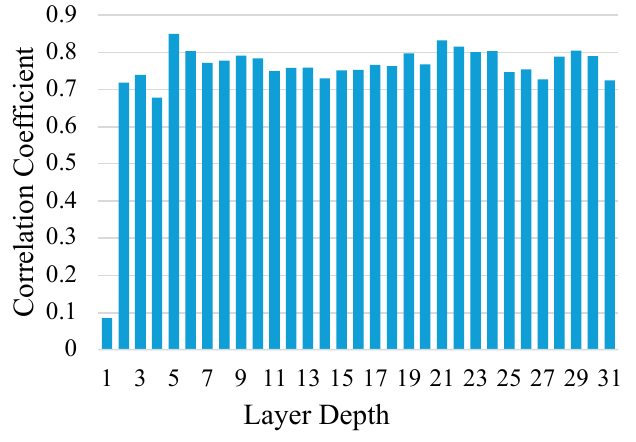}
         \caption{}
         \label{fig:correlation_scores_vs_layers}
    \end{subfigure}
    \hfill
    \begin{subfigure}[b]{0.32\textwidth}
        \includegraphics[width=\textwidth]{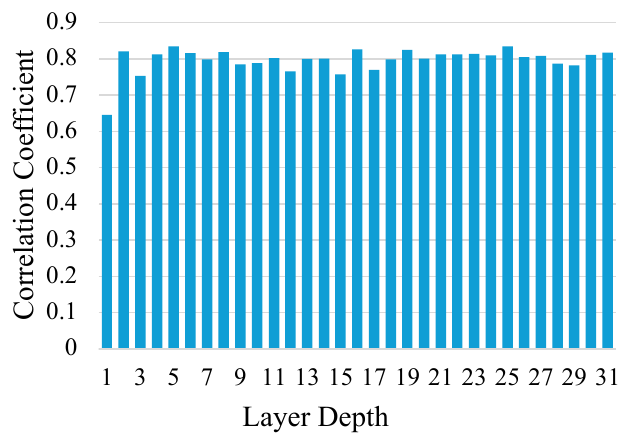}
         \caption{}
         \label{fig:correlation_scores_vs_layers_internvl2_8b}
    \end{subfigure}
    \hfill
    \begin{subfigure}[b]{0.32\textwidth}
        \includegraphics[width=\textwidth]{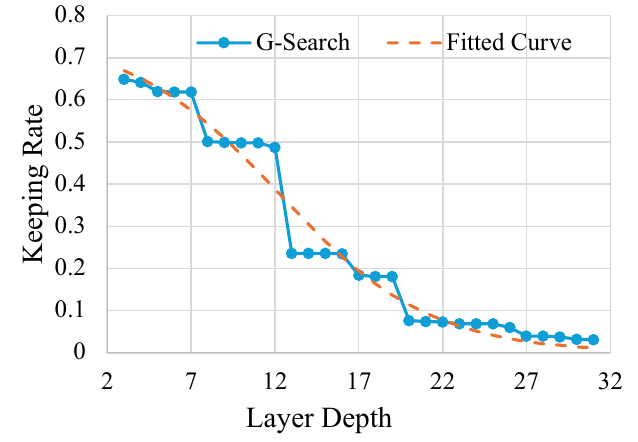}
         \caption{}
         \label{fig:keeping_rate_vs_layers}
    \end{subfigure}

    \caption{
    {\bf (a)}: Kendall's Tau correlation coefficient between the current layer and the next layer of LLaVA-1.5-7B~\cite{liu2024improved}. A value larger than 0.7 is high.
    {\bf (b)}: Kendall's Tau correlation coefficients for InternVL2-8B~\cite{chen2024internvl}.
    {\bf (c)}: Keeping rates along layers from G-Search (blue solid curve), and the fitted curve with P-Sigmoid (orange dash curve) for LLaVA-1.5-7B.
    }
    \label{fig:attn_score_corr_keep_rate}
\end{figure*}

To address the aforementioned drawbacks, in this paper, we propose a prompt-aware approach via the automatic search of optimal vision token reduction strategies for given MLLMs.
Moreover, we consider efficiency from two scenarios, \ie, (\uppercase\expandafter{\romannumeral1}) Reducing computational cost with minor performance drops, and (\uppercase\expandafter{\romannumeral2}) Improving the performance with given budgets.

We first investigate correlations between vision tokens and instruction tokens (\ie, text tokens provided by users).
Specifically, we sort the vision tokens of each layer based on their attention scores to instruction tokens and calculate the Kendall's Tau correlation coefficient~\cite{kendall1938new} between the current layer and the next layer.
Figures~\ref{fig:correlation_scores_vs_layers} and~\ref{fig:correlation_scores_vs_layers_internvl2_8b} illustrate the coefficients along layers for LLaVA-1.5-7B~\cite{liu2024improved} and InternVL2-8B~\cite{chen2024internvl}, respectively.
As shown, the correlation coefficients are high\footnote{We take coefficients over 0.7 as high based on IQA studies~\cite{zhang2011fsim,zhao2020dehazing}} since the second layer.
We regard the relative importance of one vision token as its ranking. 
Our main finding can be stated as the relative importance of one vision token remains similar in each layer of MLLMs after the first layer. 
Based on the finding, we assume that the importance of vision tokens in deeper layers can be sorted by the attention scores of earlier layers.
Concurrent works~\citep{xiao2024efficient,li2024snapkv} find that only top tokens with high attention scores are essential for LLMs' inference.
Based on our main finding, we assume that the essential top vision tokens in deeper layers are also top tokens in earlier layers.
Thus, the number of essential top tokens does not increase as the depth of the MLLM layer grows.

Based on the above analysis, we propose a greedy search (G-Search) for Scenario \uppercase\expandafter{\romannumeral1}. 
Specifically, in each layer of a pretrained MLLM from the shallow to the deep, we rank vision tokens using the attention scores of the prior layer. 
Then, we find a keeping rate to retain the least number of top vision tokens with insignificant performance drops via Bayesian Optimization. 
Our main finding and assumptions enable G-Search to remove unnecessary vision tokens for current and later layers in the current layer.
Sec.~\ref{subsec:greedy_search} shows our method can achieve the optimal vision token reduction strategy.
We analyze keeping rates from G-Search, and find that they decrease along layers and can be fitted by a S-curve as shown in Fig.~\ref{fig:keeping_rate_vs_layers}.
Accordingly, for Scenario \uppercase\expandafter{\romannumeral2}, we design a parametric sigmoid function (P-Sigmoid) to attain keeping rates along layers with given budgets, and search the optimal parameters that maximize the performance via Bayesian Optimization.

We conducted thorough experiments on 12 popular benchmarks with various MLLMs of different sizes, \ie LLaVA-1.5-7B~\citep{liu2024improved}, InternVL2 family~\citep{chen2024internvl,chen2024far} with 1B, 2B, 4B, and 8B models. 
The extensive experiments demonstrate that 
for Scenario \uppercase\expandafter{\romannumeral1}, G-Search can significantly accelerate MLLMs by up to 2.3$\times$ with only a $0.2$\% drop of average accuracy.
It scales up well, resulting in larger reduction rates for larger models.
Compared to existing prompt-aware methods, it runs $20$\% or more faster and achieves better performance.
Moreover, G-Search can further speed up prompt-agnostic methods~\citep{li2024tokenpacker,yao2024deco} by up to 1.5$\times$.
For Scenario \uppercase\expandafter{\romannumeral2}, on top of LLaVA-1.5-7B, 
P-Sigmoid outperforms FastV by +3.38\% of average accuracy when 87.5\% of vision tokens are reduced. It goes to +5.46\% with around 94\% of vision tokens reduced.
On top of InternVL2-8B, P-Sigmoid achieves a larger gain of +7.69\% when reducing 87.5\% of vision tokens.
Moreover, our results indicate that different MLLMs need different reduction strategies. Thus, handcrafted reduction methods are likely to perform well with certain MLLMs and benchmarks but fail in other cases.

Our contributions are summarized as. 
(1) We present a new insight that the relative importance of each vision token remains similar across layers of MLLMs, which paves the path for automatic reduction search. 
(2) We consider efficiency from two scenarios, and propose G-Search and P-Sigmoid that find the best reduction strategy by automatic search. G-Search significantly reduces the computational cost with minor performance drops. P-Sigmoid achieves great trade-offs between efficiency and effectiveness.
(3) We conducted extensive experiments on various benchmarks and MLLMs, and show clear boosts from the proposed methods in terms of both efficiency and efficacy.

\begin{figure*}[t]
    \centering
    \includegraphics[width=1.0\textwidth]{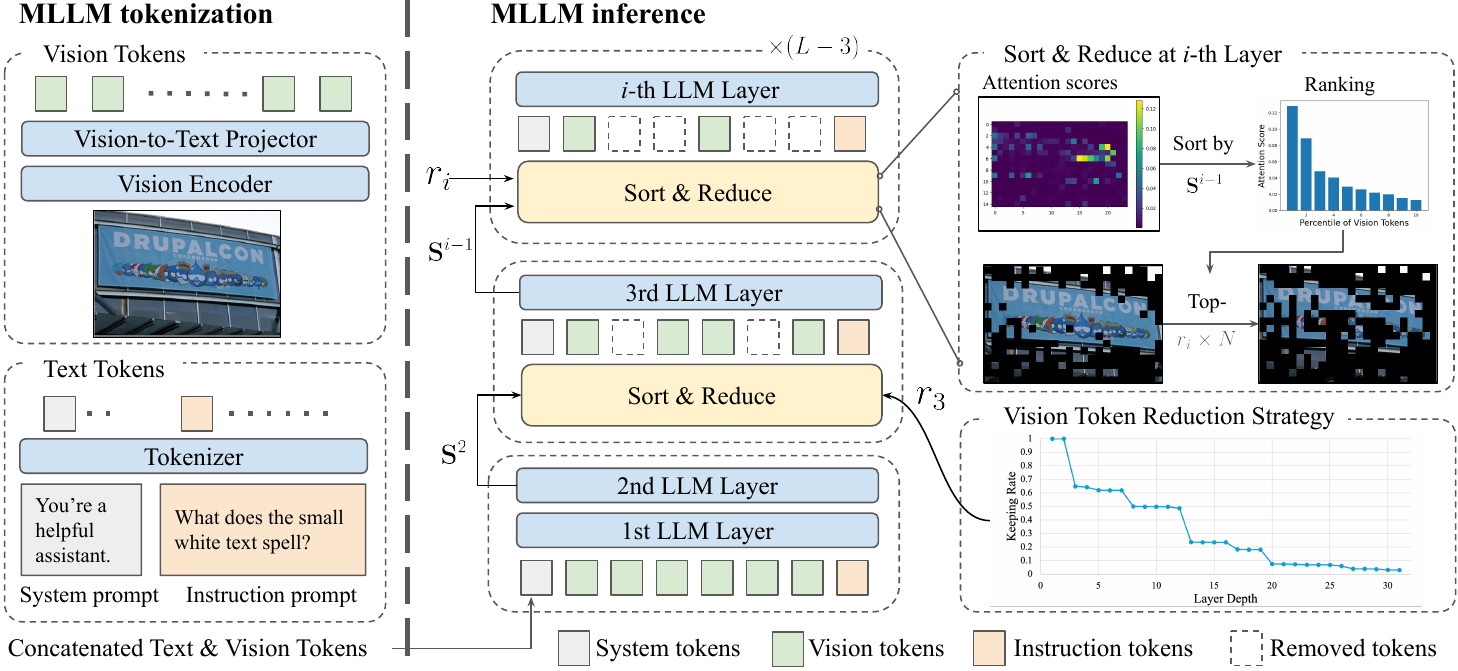}
    \caption{MLLM tokenization and inference with vision token reductions. 
    The proposed G-Search and P-Sigmoid automatically search the optimal reduction strategy, \ie the keeping rate of each layer.}
    \label{fig:lmm_data_flow}
\end{figure*}

%% file: sec/2_related_work.tex
\section{Related work}
\label{sec:related_work}

\subsection{MLLMs}

Multimodal models are able to process multiple modalities and benefit various fields~\cite{zhao2022exploiting,zhao2024taming,zhao2024generating}.
Multimodal Large Language Models (MLLMs) are LLMs that understand other modalities beyond just natural language. 
Recent focus is the integration of visual perception with language~\citep{jian2024bootstrapping,instructblip,zhu2023minigpt}. 
LLaVA~\citep{liu2023visual,liu2024improved} is one of the most powerful and popular MLLMs. It introduces linear layers to project vision features into the text embedding space, and instruction-tune both the projector and the LLM on large-scale, machine-generated instruction-following data. 
To enhance the effectiveness, later studies explore high resolution inputs~\cite{liu2024llavanext,li2024monkey}, model design~\cite{bai2023qwen}, and scaling up model size and data~\cite{chen2024internvl,chen2024far}.
Although with great performance, existing MLLMs require substantial computational resources, due to their large model sizes and the large number of vision tokens to handle.
In this paper, instead of improving the performance, we focus on improving the efficiency of MLLMs and propose training-free solutions for two scenarios.

\subsection{Token reduction}

StreamingLLM~\citep{xiao2024efficient} first shows that LLMs are likely to assign high attention scores to the first a few tokens, called attention sinks. 
With attention sinks and a fixed-length attention window in the key-value (KV) caches, LLMs are able to achieves the similar performance as the full KV caches are used.
Later, SnapKV~\citep{li2024snapkv} automatically compresses KV caches by selecting clustered important KV positions for each attention head.
Those methods reduce computational cost of the decoding phrase but still calculate the whole KV caches in the prefilling phrase.
In contrast, our methods can reduce the cost of both prefilling and decoding phrases.

Besides text token reduction in LLM, recent studies also explore vision token reduction in MLLMs.
LLaVA-HR~\citep{luo2024feast} and Mini-Gemini~\citep{li2024mini} fuse vision tokens of high resolution inputs into low resolution ones.
LLaVA-UHD~\citep{xu2024llava} and Beyond LLaVA-HD~\citep{zhang2024beyond} learns local compression layers.
LLaMA-PruMerge~\citep{shang2024llava} removes vision tokens from a ViT model~\cite{dosovitskiy2021an}, which are not highly correlated to the class token.
Honeybee~\citep{cha2024honeybee}, TokenPacker~\citep{li2024tokenpacker}, and Deco~\citep{yao2024deco} leverage convolutions, local attentions, and average pooling to down-sample the features of high resolution images.
Those methods are prompt-agnostic, which reduce vision tokens before feeding them into the LLM. Thus, they ignore the user instructions that are important clues to remove irrelevant vision tokens.
Prompt-aware approaches consider user instructions.
LLaMA-VID~\citep{li2024llamavid} and VoCo-LLaMA~\citep{ye2024voco} compress vision tokens into a few tokens based on vision-text cross attentions. However, their compression comes with a clear performance drop.
FastV~\citep{chen2024image}, VTW~\citep{lin2024boosting}, and PDrop~\cite{xing2024pyramiddrop} reduce vision tokens at certain layers of LLM based on the attention scores between vision and instructions tokens.
However, their reduction strategy is manually crafted based on a certain MLLM or a benchmark.
In contrast, our method automatically finds the optimal reduction strategy, generalizes to various MLLMs and benchmarks, and achieves better efficiency-effectiveness trade-off.

%% file: sec/3_method.tex
\section{Searching optimal reduction strategy}
\label{sec:method}

\begin{algorithm}[tb]
    \caption{G-Search with Bayesian Optimization}\label{alg:greedy_search}
    {\bf Input:}
\begin{algorithmic}
    \State Target Function $f(\cdot)$,
    \State MLLM $\theta$, 
    \State Data Samples $\mathcal{D}$, 
    \State Number of Bayesian Optimization Iterations $T$
\end{algorithmic}
    {\bf Output:} %
    Keeping Rate Sequence $\mathcal{R}$ 
\begin{algorithmic}
    \LineComment{Define a function for Bayesian Optimization}
    \Function {BayesianOptimization}{$f$, $\mathcal{R}$, $\theta$, $\mathcal{D}$, $T$}
        \State Initialize a Gaussian Process model $\mathcal{GP}$
        \State Define the acquisition function $\mathrm{A}(\cdot)$
        \LineComment{Expected Improvement (EI) is adopted as $\mathrm{A}(\cdot)$}
        \State Uniformly sample $\mathcal{X}_0 = \{\hat{r}_{i<10} | \hat{r}_i \in [0, min(\mathcal{R})]$\}
        \State $\forall \hat{r} \in \mathcal{X}_0$, evaluate the target function $f(\hat{r}|\mathcal{R}, \theta, \mathcal{D})$.
        \For{$n = 0$ to $T-1$}
            \State Fit $\mathcal{GP}$ to the observed data $(\mathcal{X}_n, f(\mathcal{X}_n|\mathcal{R}, \theta, \mathcal{D}))$.
            \State Get the next point $\hat{r}_{n+1} = \arg\max\limits_{r} \mathrm{A}(r; \mathcal{GP})$
            \State Update $\mathcal{X}_{n+1} = \mathcal{X}_n \cup \{\hat{r}_{n+1}\}$
        \EndFor
        \State Find the point w/ the best observed  value by Eq.~\ref{eq:target_f}
        \State \Return{$r^*$}
    \EndFunction
    
    \LineComment{Repeat searching for each layer}
    
    \State $\mathcal{R} \gets \emptyset$
    \For{layer depth $i \in [3,..,L]$}
        \State $r_i$ = \Call{BayesianOptimization}{$f$, $\mathcal{R}$, $\theta$, $\mathcal{D}$, $T$}
        \State $\mathcal{R} \gets \mathcal{R} \cup r_i$
    \EndFor

\end{algorithmic}
\end{algorithm}

\subsection{Correlation of vision \& instruction tokens}
\label{subsec:relationship_vision_instruction}

{\bf Preliminaries:} 
Prevailing MLLM architecture comprises a vision encoder, a vision-to-text projector, and a pre-trained LLM.
The vision encoder is usually a pre-trained vision transformer of CLIP~\citep{radford2021learning} that encodes an image into vision tokens.
The vision-to-text projector projects vision tokens of the vision encoder into the text space. 
Q-former~\citep{li2023blip} and linear layers~\citep{liu2024improved} are popular choices for the projector.
The pre-trained LLM takes as input system prompt tokens, projected vision tokens, and instruction tokens, and generates responses. Usually, system tokens remain unchanged for a given LLM, and instruction tokens vary as the user inputs. 
We visualize the tokenization of MLLMs and how to feed tokens into the LLM in the left part of Fig.~\ref{fig:lmm_data_flow}.

\noindent
{\bf Why cross-modality token correlation matters:} 
Different user instructions usually refer to different regions of images or vision tokens, leaving others irrelevant. 
In this paper, we attempt to leverage instruction tokens as a clue to remove irrelevant vision tokens, thus accelerating MLLMs.
Since attention scores are widely used to interpret the alignment between tokens~\citep{chefer2021transformer,xiao2024efficient,li2024snapkv}, we leverage attention scores between vision tokens and instruction tokens to indicate their correlation.
Vision tokens of high attention scores are highly correlated to the user instructions and thus are important tokens to keep.
We run experiments on a holdout dataset of LLaVA~\cite{liu2024improved} and have the following findings.

\noindent
{\bf Main finding:} 
\emph{The ranking or relative importance of each vision token remains similar in each layer after the first layer.} 
In each layer of the model, we rank the vision tokens based on their attention scores and calculate the Kendall's Tau correlation coefficient~\cite{kendall1938new} between the ranking of the current layer and that of the next layer.
Figures~\ref{fig:correlation_scores_vs_layers} and ~\ref{fig:correlation_scores_vs_layers_internvl2_8b} visualize the correlation coefficients along layers for LLaVA-1.5-7B and InternVL2-8B, respectively.
As shown, the correlation coefficients are high (usually $\geq0.7$) except the one between the first two layers, which indicate the relative importance of each vision token remains similar across layers except the first layer.

\subsection{G-Search for Scenario \uppercase\expandafter{\romannumeral1}}
\label{subsec:greedy_search}

\noindent
{\bf Assumption (1)}: 
\emph{The importance of vision tokens in deeper layers can be decided by attention scores of earlier layers.}
This is directly derived from our main finding in Sec.~\ref{subsec:relationship_vision_instruction} that the ranking of vision tokens is similar across layers. 

\noindent
{\bf Assumption (2)}: 
\emph{The number of essential top vision tokens does not increase along layers.}
This assumption is derived by the following.
First, recent studies~\citep{xiao2024efficient,li2024snapkv} demonstrate that only top tokens with high attention scores are essential for LLMs' inference.
According to our main finding, those essential top tokens in deeper layers should be top tokens in earlier layers.
Second, if some tokens should be kept in deeper layers to preserve the performance, we cannot remove them from earlier layers. Otherwise, deeper layers cannot access those tokens. Thus, essential tokens in deeper layers are a subset of essential tokens in earlier layers.

\noindent
{\bf Greedy Search Algorithm (G-Search):}
Based on our assumptions, we propose the following greedy algorithm to find the least number of essential tokens at each layer, and prove that it can reach the optimal solution.

For the $i$-th layer $i=3,..,L$ of a pretrained MLLM, we sort vision tokens based on their attention scores ($\mathbf{S}^{i-1}$) of the last layer, \ie the $(i-1)$-th layer.
Note that we ignore the first layer because it is poorly correlated with other layers.
Then, we search a keeping rate $r_i \in [0,1]$ that decides the number of top vision tokens ($n_i$) to keep at the $i$-th layer w.r.t. the total number of input vision tokens ($N$). 
That is, $r_i = {n_i}/{N}$.
We regard $r_i^*$ as the optimal keeping rate that maximizes the target function $f(\cdot)$, 
\begin{align} \label{eq:target_f}
    r_i^* & = \underset{r_i \leq r_{i-1}}{\mathrm{argmax}} f(r_i) \\
    f(r_i) & = E(r_i|r_3,r_4, ..., r_{i-1}, \theta, D) - \lambda \cdot r_i
\end{align}
where $E(\cdot)$ refers to the performance of the MLLM parameterized by $\theta$ on the dataset $D$. The term $E(\cdot)$ refers to effectiveness, and the term with $r_i$ embodies efficiency. 
We set $\lambda=0.01$ so that efficiency is improved when performance is maintained.
Since ${n_i}$ and $N$ are bounded integers, we can always get the optimal $r_i^*$ via brute force. 
For a faster search, we employ Bayesian Optimization~\citep{mockus2005bayesian}.
The pseudo implementation is provided in Algorithm~\ref{alg:greedy_search}.

\noindent
{\bf Inference with keeping rates:} 
As shown in Fig.~\ref{fig:lmm_data_flow}, the only different between the standard inference and inference with keeping rates is to add a plug-and-play Sort \& Reduce module before each LLM layer.
In Sort \& Reduce of the $i$-th layer, similar as G-Search, we first sort the vision tokens based on their attention scores of the $(i-1)$-th layer. Then, we keep the top $(r_i\cdot N)$ tokens and remove the rest.

\noindent
{\bf Proof of the optimality:}
Supposing the optimal sequence of keeping rates is $\mathcal{R}^* = {[r_3^*,r_4^*, ..., r_L^*]}$, the sequence from our search is $\mathcal{R} = {[r_3,r_4, ..., r_L]}$, and $\forall i<j, r_i=r_i^*$. 

If $r_j^* > r_j$, our method reduces more tokens (noted as $v^+$) than the optimal at the $j$-th layer. 
That is, $v^+$ are not necessary in current layer and can also be removed in later layers. Otherwise, if later layers require $v^+$, the current layer also needs $v^+$ based on our Assumption (2).
Thus, we can replace $r_j^*$ with $r_j$ from our search.

If $r_j^* < r_j$, our method reduces less tokens than the optimal at the $j$-th layer. Since our method always remains the performance, $r_j^*$ and $r_j$ results in the same performance. Based on our greedy strategy, we should find $r_j^*$ as $r_j$. Thus, $r_j^* < r_j$ is impossible.

In conclusion, we can always convert $\mathcal{R}^*$ to $\mathcal{R}$, showing that $\mathcal{R}$ from our search is optimal.

\begin{figure}[tb]
    \centering
    \includegraphics[width=0.95\columnwidth]{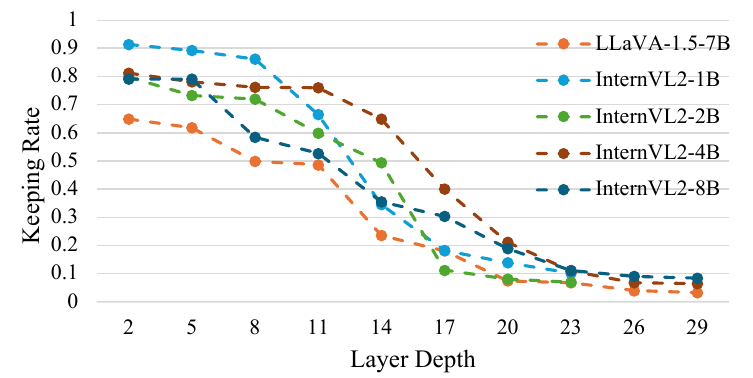}
    \caption{Keeping rates of various MLLMs from G-Search. All curves are S-curve and can be fitted by P-Sigmoid.}
    \label{fig:keep_rates_from_gsearch}
\end{figure}

\subsection{P-Sigmoid for Scenario~\uppercase\expandafter{\romannumeral2}}
Scenario~\uppercase\expandafter{\romannumeral2} requires improving the performance with a given budget.
We start with analyzing the reduction strategies of G-Search and find that keeping rates along layers can be fitted into a sigmoid-like curve as shown in Fig.~\ref{fig:keeping_rate_vs_layers}.
Keeping rates of different MLLMs are visualized in Fig.~\ref{fig:keep_rates_from_gsearch}, which are all in sigmoid-like curves.
We call the the sigmoid-like function to fit as the parametric sigmoid (P-Sigmoid) and define it as,
\begin{align}
    \hat{r}(i) = \frac{2b}{1+e^{k(i-\alpha)}}
\end{align}
where $\hat{r}(i)$ is the fitted keeping rate for the $i$-th layer, and $b \in [0,1]$ refers to the rate of the numbers of vision tokens after and before the reduction.
We take $b$ as the budget because computational cost is positively correlated to the number of tokens.
$\alpha$ is the midpoint of the domain of $\hat{r}(\cdot)$. 
For example, the function $\hat{r}(\cdot)$ for a MLLM has the domain of $[3, L]$, and $\alpha = 17.5$. Note that we do not reduce vision tokens of the first two layers.

The integral of $\hat{r}(\cdot)$ within its domain of $[3, L]$ is always $(L-2)*b$ regardless of $k$, which enables vision token reduction at a given budget. 
We assume that, in Scenario \uppercase\expandafter{\romannumeral2}, the optimal keeping rates follow similar parametric sigmoid functions as $\hat{r}(\cdot)$.
Since $b$ and $\alpha$ are known for a specific MLLM, we search the optimal $k$ to maximize the performance on a small dataset.
We leverage Bayesian Optimization to search a non-negative real number for $k$.

%% file: sec/4_experiments.tex
\section{Experiments}
\label{sec:experiment}

\subsection{Experimental setup}

{\bf Benchmarks:} 
We adopt 12 popular evaluation benchmarks and follow Cambrian-1~\cite{tong2024cambrian} to categorize them as,
\begin{itemize}
    \item \emph{General VQA}: MME, MMBench~\cite{liu2025mmbench}, and GQA~\cite{hudson2019gqa}. Those benchmarks include a wide range of visual question answering questions and indicate the general ability of MLLMs on visual understanding.
     \item \emph{Knowledge}: MMMU~\cite{yue2024mmmu}, MathVista~\cite{lu2023mathvista}, and AI2D~\cite{kembhavi2016diagram}. They mainly focus on knowledge test across disciplines, including Art, Business, Health \& Medicine,, Humanities, Math, and Tech \& Engineering.
     \item \emph{OCR \& Chart}: TextVQA~\cite{singh2019towards}, ChartVQA~\cite{masry-etal-2022-chartqa}, and DocVQA~\cite{mathew2021docvqa}, which focus on the understanding of charts, diagrams, and documents. 
     \item \emph{Vision-Centric}: POPE~\cite{li2023evaluating}, RealWorldQA~\cite{RealWordQA}, and HallusionBench~\cite{guan2024hallusionbench}. They are used to evaluate MLLMs in terms of language hallucination and visual illusion.
\end{itemize}

\noindent
{\bf Metrics for effectiveness:}
We adopt default metrics of each benchmark.
Specifically, the accuracy is the metric for most benchmarks. 
And we report the relaxed accuracy~\cite{masry-etal-2022-chartqa} for ChartQA, Average Normalized Levenshtein Similarity (ANLS)~\cite{biten2019scene} for DocVQA, F1 scores for POPE, and the sum of perception and recognition scores for MME.
When calculating the average accuracy, we normalize MME scores by dividing the full score, \ie 2800.

\noindent
{\bf Metrics for efficiency:}
We evaluate the efficiency in terms of memory cost, FLOPs, and time cost.
For the memory cost, we mainly consider the KV-Cache of vision tokens and report the rate of the numbers of kept vision tokens and vision tokens without reduction.
To calculate FLOPs, we employ the tool calflops~\cite{calflops} and report Tera FLOPs (TFLOPs).
For the time cost, we use optimum-benchmark~\footnote{https://github.com/huggingface/optimum-benchmark} from Huggingface to evaluate the pre-filling time at inference.
This tool requires to preset the number of input text tokens and the number of output tokens. 
To mimic the evaluation on 12 benchmarks, we calculate the mean numbers of input and output text tokens on all samples. It ends up 75 input tokens and 5 output tokens after rounding.

\noindent
{\bf MLLMs in experiments:}
In our experiments, we compare our methods and existing reduction methods on top of various MLLMs, including the popular LLaVA-1.5-7B, and InternVL2-1B/2B/4B/8B from the InternVL family~\citep{chen2024internvl,chen2024far} that are one of SOTA open sourced MLLMs.

\begin{table}[tb]
\centering
\setlength\tabcolsep{4.5pt} 
\resizebox{\columnwidth}{!}{  
    \begin{tabular}{l|cccc}
    \toprule
    \multirow{2}{*}{\shortstack{MLLM \ \ \ \ \\+ Method}} & Avg. & Memory  & \multirow{2}{*}{TFLOPs $\downarrow$} & Time   \\
           & acc. $\uparrow$ & cost $\downarrow$ & ~ & cost $\downarrow$  \\
    \hline
    LLaVA-1.5-7B & 48.97 & 1.0 & 9.18 &  0.625  \\
    + VTW & 44.32 & 0.5 & 5.19 &  0.385 \\
    + PDrop & \underline{48.70} & \underline{0.469} & \underline{5.47} & \underline{0.381} \\
    + FastV (R=50\%) & \underline{48.70} & 0.531 & 5.49 &  0.387  \\
    + G-Search (Ours) & {\bf 48.77} & {\bf 0.340} & {\bf 3.95} &  {\bf 0.301}  \\

    \hline
    InternVL2-1B & 59.85 & 1.0 & 4.62 &  0.384  \\
    + VTW & 41.13 & {\bf 0.5} & {\bf 3.73} & {\bf 0.331} \\
    + PDrop & 53.70 & 0.583 & 3.88 & 0.336 \\
    + FastV (R=50\%) & \underline{54.85} & 0.542 & 3.91 &  0.342  \\
    + G-Search (Ours) & {\bf 59.19} & \underline{0.527} & \underline{3.84} & \underline{0.333}  \\

    \hline
    InternVL2-2B & 61.94 & 1.0 & 8.10 & 0.598  \\
    + VTW & 37.84 & {\bf 0.5} & {\bf 5.50} & {\bf 0.439}  \\
    + PDrop & 58.98 & 0.583 & 5.93 & 0.452 \\
    + FastV (R=50\%) & 59.91 & 0.542 & 5.85 & 0.451  \\
    + G-Search (Ours) & {\bf 61.22} & \underline{0.532} & \underline{5.64} &  \underline{0.444}  \\

    \hline
    InternVL2-4B & 68.16 & 1.0 & 13.97 &  0.969  \\
    + VTW & 49.01 & 0.5 & 8.72 &  0.649  \\
    + PDrop & \underline{66.94} & {\bf 0.469} & {\bf 8.35} & {\bf 0.627} \\
    + FastV (R=50\%) & 66.19 & 0.531 & 9.01 &  0.652  \\
    + G-Search (Ours) & {\bf 67.65} & \underline{0.488} & \underline{8.69} &  \underline{0.645}  \\
    
    \hline
    InternVL2-8B & 70.83 & 1.0 & 24.10 &  1.518  \\
    + VTW & 52.51 & 0.5 & 13.71 & 0.927  \\
    + PDrop & 69.19 & \underline{0.469} & \underline{13.13} & \underline{0.915} \\
    + FastV (R=50\%) & \underline{69.42} & 0.531 & 14.58 & 0.998 \\
    + G-Search (Ours) & {\bf 70.10} & {\bf 0.424} & {\bf 12.24} & {\bf 0.860}  \\
    
    \bottomrule
    \end{tabular}
}
\caption{
Comparison to prompt-aware reduction methods for Scenario~\uppercase\expandafter{\romannumeral1}.
``Avg. acc.'' refers to average accuracy on 12 benchmarks.
The top-2 best results among those methods are highlighted in bold and underline, respectively.
}
\label{table:run_time_eval_s1}
\end{table}

\noindent
{\bf Implementation details:}
LMMs-Eval~\cite{zhang2024lmmsevalrealitycheckevaluation} was adopted to evaluate MLLMs on the 12 aforementioned benchmarks. 
Note that for LLaVA-1.5-7B, a lower performance on TextVQA is expected because the official evaluation code adds extra reference OCR tokens for inference, which allows MLLMs to make choices between possible answers instead of understanding vision tokens.
It is not necessary to search at each layer, as you can see flat regions in Fig.~\ref{fig:keeping_rate_vs_layers}. Thus, we conducted G-Search for every three layers.
All searches were conducted on a small split of training data of LLaVA~\cite{liu2024improved}.
Since our methods are training-free, all experiments can be conducted using 8 NVIDIA A100 GPUs.

\begin{table}[tb]
\centering
\setlength\tabcolsep{4.5pt} 
\resizebox{\columnwidth}{!}{  
    \begin{tabular}{l|cccc}
    \toprule
    \multirow{2}{*}{\shortstack{LLaVA-1.5-7B\\+ Method \ \ \ \ \ \ \ \ }} & Avg. & Memory  & \multirow{2}{*}{TFLOPs $\downarrow$} & Time   \\
           & acc. $\uparrow$ & cost $\downarrow$ & ~ & cost $\downarrow$  \\
    \hline
    TokenPacker & 47.60 & 1.0 & 3.27 & 0.268 \\
    + G-Search (Ours) & 47.68 & 0.411 & 2.12 & 0.182 \\
    \multicolumn{1}{c|}{$\Delta$} & {\bf +0.08} & {\bf -0.589} & {\bf -1.15} & {\bf -0.086} \\
    \hline
    DeCo & 46.97 & 1.0 & 3.26 & 0.268  \\
    + G-Search (Ours) & 46.71 & 0.432 & 2.16 & 0.198 \\
    \multicolumn{1}{c|}{$\Delta$} & -0.26 & {\bf -0.568} & {\bf -1.1} & {\bf -0.070} \\

    \bottomrule
    \end{tabular}
}
\caption{
G-Search improves prompt-agnostic reduction methods.
}
\label{table:run_time_eval_on_prompt_agnostic}
\end{table}

\begin{table*}[t]
\centering
\setlength\tabcolsep{3.5pt} 
\resizebox{\textwidth}{!}{
    \begin{tabular}{cl|rr|rrr|rrr|rrr|rrr}
    \toprule
      MLLM & Method & \multicolumn{1}{c}{} &  \multicolumn{1}{c|}{} & \multicolumn{3}{c|}{General VQA} & \multicolumn{3}{c|}{Knowledge} & \multicolumn{3}{c|}{OCR \& Chart} & \multicolumn{3}{c}{Vision-Centric}  \\
      ~ & ~ & \multicolumn{1}{c}{\rotatebox{90}{TFLOPs $\leftarrow$}} & \multicolumn{1}{c|}{\rotatebox{90}{Avg acc. $\rightarrow$}} & \multicolumn{1}{c}{\rotatebox{90}{MME}} & \multicolumn{1}{c}{\rotatebox{90}{MMBench}} & \multicolumn{1}{c|}{\rotatebox{90}{GQA}} & \multicolumn{1}{c}{\rotatebox{90}{MMMU}} & \multicolumn{1}{c}{\rotatebox{90}{MathVista}} & \multicolumn{1}{c|}{\rotatebox{90}{AI2D}} & \multicolumn{1}{c}{\rotatebox{90}{TextVQA}} & \multicolumn{1}{c}{\rotatebox{90}{ChartQA}} & \multicolumn{1}{c|}{\rotatebox{90}{DocVQA}} & \multicolumn{1}{c}{\rotatebox{90}{POPE}} &  \multicolumn{1}{c}{\rotatebox{90}{RealWorldQA}} & \multicolumn{1}{c}{\rotatebox{90}{HallusionBench}} \\
      \hline

    \multirow{3}{*}{\shortstack{LLaVA\\-1.5-7B}} 
    & + FastV & 2.74 & 43.14 & 1644.5 & 60.2 & 54.0 & 37.3 & 21.8 & 52.1 & 38.7 & 14.4 & 18.1 & 66.3 & 49.3 & 46.7 \\
    ~ & + P-Sigmoid & 2.66 & 46.52 & 1700.4 & 63.7 & 58.3 & 37.1 & 22.0 & 54.5 & 44.1 & 15.7 & 23.3 & 78.0 & 53.6 & 47.1 \\
    ~ & \multicolumn{1}{c|}{$\Delta$} & {\bf -0.08} & {\bf +3.38} & {\bf +55.9} & {\bf +3.5} & {\bf +4.3} & -0.2 & {\bf +0.2} & {\bf +2.4} & {\bf +5.4} & {\bf +1.3} & {\bf +5.2} & {\bf +11.7} & {\bf +4.3} & {\bf +0.4} \\
    \hline

    \multirow{3}{*}{\shortstack{Intern\\VL2-1B}} 
    & + FastV & 3.43 & 43.16 & 1581.8 & 52.5 & 47.5 & 34.0 & 23.8 & 55.0 & 39.6 & 17.2 & 26.5 & 79.4 & 43.3 & 42.8 \\
    & + P-Sigmoid & 3.38 & 47.83 & 1691.7 & 56.6 & 50.8 & 33.3 & 26.6 & 55.5 & 49.6 & 33.9 & 36.2 & 83.2 & 44.8 & 42.9 \\
    & \multicolumn{1}{c|}{$\Delta$} & {\bf -0.05} & {\bf +4.67} & {\bf +109.9} & {\bf +4.1} & {\bf +3.3} & -0.7 & {\bf +2.8} & {\bf +0.5} & {\bf +10.0} & {\bf +16.7} & {\bf +9.7} & {\bf +3.8} & {\bf +1.5} & {\bf +0.1} \\
    \hline

    \multirow{3}{*}{\shortstack{Intern\\VL2-2B}} 
    & + FastV & 4.26 & 47.66 & 1661.0 & 68.1 & 52.8 & 33.3 & 25.9 & 68.0 & 54.5 & 23.8 & 34.0 & 80.1 & 29.7 & 42.3 \\
    & + P-Sigmoid & 4.16 & 51.54 & 1710.7 & 68.1 & 55.0 & 33.2 & 28.6 & 66.7 & 61.3 & 43.0 & 46.5 & 82.2 & 30.1 & 42.8 \\
    & \multicolumn{1}{c|}{$\Delta$} & {\bf -0.10} & {\bf +3.88} & {\bf +49.7} & 0.0 & {\bf +2.2} & -0.1 & {\bf +2.7} & -1.3 & {\bf +6.8} & {\bf +19.2} & {\bf +12.5} & {\bf +2.1} & {\bf +0.4} & {\bf +0.5} \\
    \hline

    \multirow{3}{*}{\shortstack{Intern\\VL2-4B}} 
    & + FastV & 5.48 & 54.83 & 1950.9 & 74.1 & 56.7 & 44.1 & 25.8 & 70.9 & 57.3 & 34.1 & 42.0 & 81.6 & 55.8 & 45.9 \\
    & + P-Sigmoid & 5.38 & 61.19 & 2039.1 & 76.3 & 59.4 & 45.8 & 30.0 & 74.3 & 66.1 & 60.8 & 57.3 & 84.0 & 57.8 & 49.7 \\
    & \multicolumn{1}{c|}{$\Delta$} & {\bf -0.10} & {\bf +6.36} & {\bf +88.2} & {\bf +2.2} & {\bf +2.7} & {\bf +1.7} & {\bf +4.2} & {\bf +3.4} & {\bf +8.8} & {\bf +26.7} & {\bf +15.3} & {\bf +2.4} & {\bf +2.0} & {\bf +3.8} \\
    \hline
    
    \multirow{3}{*}{\shortstack{Intern\\VL2-8B}} 
    & + FastV & 7.71 & 55.17 & 2066.2 & 74.3 & 55.4 & 44.7 & 25.6 & 72.6 & 60.3 & 35.0 & 39.6 & 78.9 & 54.5 & 47.3 \\
    & + P-Sigmoid & 7.46 & 62.86 & 2176.6 & 79.3 & 59.7 & 45.8 & 33.0 & 76.3 & 70.5 & 61.8 & 59.9 & 84.1 & 59.5 & 46.7 \\
    & \multicolumn{1}{c|}{$\Delta$} & {\bf -0.25} & {\bf +7.69} & {\bf +110.4} & {\bf +5.0} & {\bf +4.3} & {\bf +1.1} & {\bf +7.4} & {\bf +3.7} & {\bf +10.2} & {\bf +26.8} & {\bf +20.3} & {\bf +5.2} & {\bf +5.0} & -0.6 \\
    
    \bottomrule
    \end{tabular}
}
\caption{P-Sigmoid with different MLLMs. We set R=87.5\% for FastV and set the budget of P-Sigmoid close to that of FastV.}
\label{table:p_sigmoid_diff_mllm}
\end{table*}

\subsection{Evaluation for Scenario \uppercase\expandafter{\romannumeral1}}
\label{subsec:eval_scenario_1}

For Scenario \uppercase\expandafter{\romannumeral1}, we attempt to accelerate the MLLMs without significant performance drops. 
We first compare our G-Search with existing prompt-aware methods, \ie, VTW~\citep{lin2024boosting}, PDrop~\cite{xing2024pyramiddrop}, and FastV~\citep{chen2024image}.
Then, we demonstrate that G-Search can further improve the efficiency on top of two recent prompt-agnostic methods that reduce vision tokens before feeding them into the LLM., \ie TokenPacker~\citep{li2024tokenpacker}, and DeCo~\citep{yao2024deco}.

\noindent
{\bf Comparison to existing prompt-aware methods:} 
We report the main results in Table~\ref{table:run_time_eval_s1} with the following findings. Please check the supplement for full results on all benchmarks.
First, our G-Search reduces TFLOPs by 16.9\%, 30.4\%, 35.9\%, and 49.2\% on top of InternVL2-1B, 2B, 4B, and 8B, respectively, achieving larger reduction rates on larger models. This is probably because larger models are likely to have more computational redundancy.
Second, compared to VTW~\citep{lin2024boosting}, PDrop~\cite{xing2024pyramiddrop}, and FastV~\citep{chen2024image}, our G-Search either has much lower TFLOPs or achieves better average accuracy.
For example, on top of LLaVA-1.5-7B, our method gets similar average accuracy as FastV with only $72\%$ TFLOPs.
Third, on top of InternVL2 models, prior methods hardly maintain the performance as they do on LLaVA-1.5-7B. 
Since their reduction strategy is manually designed based on LLaVA-1.5-7B and specific benchmarks, their hand crafted reductions do not generalize to different MLLMs and benchmarks.
In contrast, thanks to the automatic search, G-Search generalizes well to different MLLMs and benchmarks.

\noindent
{\bf Improving prompt-agnostic methods:}
We apply our G-Search on top of two recent prompt-agnostic methods, \ie TokenPacker~\citep{li2024tokenpacker}, and DeCo~\citep{yao2024deco}.
As shown in Table~\ref{table:run_time_eval_on_prompt_agnostic}, our method consistently reduces TFLOPs by more than 33.9\% with less than 0.21\% drop on the average accuracy.
TokenPacker and DeCo reduce vision tokens without considering user instructions.
In contrast, our method is aware of user instructions and improves them probably by removing more irrelevant vision tokens.

\begin{figure}[t]
    \centering
    \includegraphics[width=1.0\columnwidth]{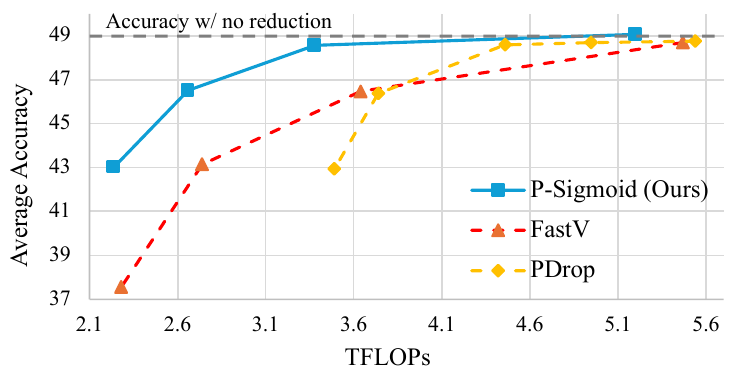}
    \vspace{-8mm}
    \caption{Comparison of reduction methods with different budgets. The average accuracy of LLaVA-1.5-7B with reductions on 12 benchmarks are reported. Compared to others, Our P-Sigmoid achieves the accuracy with no reduction using less TFLOPs.}
    \label{fig:diff_budget}
\end{figure}

\subsection{Evaluation for Scenario \uppercase\expandafter{\romannumeral2}}
\label{subsec:eval_scenario_2}

\noindent
{\bf P-Sigmoid with different budgets:}
FastV is configurable with a filtering ratio (R) that refers to the percentage of vision tokens to remove at the 2nd layer.
We instantiate FastV with different R to set up different budgets.
The values of R are set as $50\%$, 75\% (\ie 25\% vision tokens left), $87.5\%$, $93.75\%$. 
Then, we set $b$ of P-Sigmoid to match the budgets of FastV.
For P-Sigmoid, the number of vision tokens are different across layers, which requires slightly more TFLOPs than keeping the same number of vision tokens in every layer as FastV. See theoretical analysis in the supplement.
Thus, to get close TFLOPs as FastV, we lower down the budgets for our method by $\sim$7\%.
For PDrop, we use different $\lambda$ to set up different budgets.
Our P-Sigmoid is compared with FastV and PDrop on top of LLaVA-1.5-7B in Fig.~\ref{fig:diff_budget}.
As shown, P-Sigmoid outperforms both FastV and PDrop in a large margin when TFLOPs are low, which clearly demonstrates that our method achieves a better trade-off between effectiveness and efficiency.

\noindent
{\bf P-Sigmoid with different MLLMs:}
We set R=87.5\% for FastV, and compare it with the proposed P-Sigmoid on top of different MLLMs.
As shown in Table~\ref{table:p_sigmoid_diff_mllm}, with similar TFLOPs, our method outperforms FastV by a large margin on top of all MLLMs, \ie, +3.38\% for LLaVA-1.5-7B, and +4.67\%/+3.88\%/+6.36\%/+7.69\% for InternVL2-1B/2B/4B/8B.
Moreover, our method gets the most improvement on OCR \& Chart and Vision-Centric benchmarks, and gets moderate gains on Knowledge benchmarks.
For example, on top of InternVL2-8B, P-Sigmoid and FastV get the scores of 61.8 vs 35.0 on ChartQA, while they get 45.8 vs 44.7 on MMMU.
Since ChartQA is more about visual recognition and understanding, our P-Sigmoid significantly outperforms FastV by providing a better way to preserve vision tokens.
MMMU focuses on knowledge test where texts are usually sufficient to address the questions. Thus, how to reduce vision tokens matters less.

\subsection{Further Analysis}

\begin{figure}[t]
    \centering
    \includegraphics[width=1.0\columnwidth]{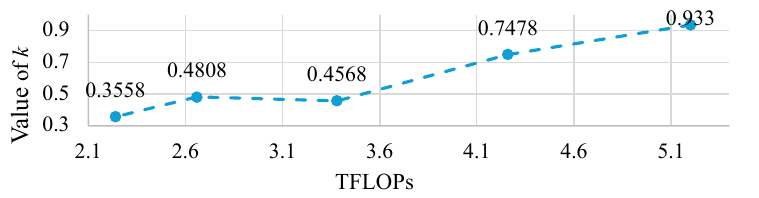}
    \caption{Values of $k$ of P-Sigmoid vs TFLOPs.}
    \label{fig:k_vs_tflops_llava_7b}
\end{figure}

\noindent
{\bf Smaller $k$ of P-Sigmoid for lower budgets:}
As shown in Fig.~\ref{fig:k_vs_tflops_llava_7b}, on top of LLaVA-1.5-7B, the value of $k$ of P-Sigmoid increases as TFLOPs increase.
A larger $k$ means a sharper S-curve, and a smaller $k$ refers to a flatter curve. That is, we are assigning more percentages of the budget to early layers when the budget increases.
A possible explanation is that there are a few essential tokens in deep layers. Without those tokens, the performance will significantly drop. 
As a result, when the budget is limited, we need to assign enough computations to deep layers for those essential tokens.
When the budget increases, deep layers get enough computations for essential tokens, and the rest of the budget can be assigned to early layers.

\begin{figure}[t]
\centering
     \includegraphics[width=\columnwidth]{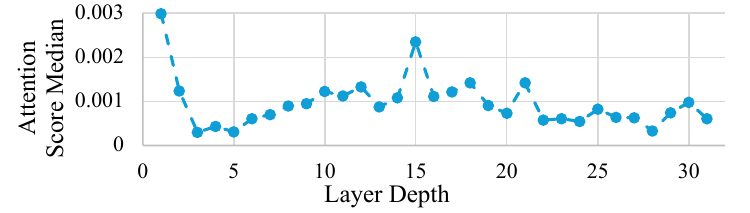}
    \caption{
     Attention scores vs layers of LLaVA-1.5-7B.
     }
     \label{fig:attn_scres_llava_7b_small}
\end{figure}

\begin{table}[tb]
\centering
    \begin{tabular}{lc|c}
    \toprule
    \multirow{2}{*}{MLLM} & Reduction & Average   \\
           & strategy of & accuracy $\uparrow$   \\
    \hline
    InternVL2-1B & Self & {\bf 59.19}  \\
    InternVL2-1B & InternVL2-2B & 57.58  \\
    \hline
    InternVL2-2B & Self & {\bf 61.22}    \\
    InternVL2-2B & InternVL2-2B & 60.90   \\
    \hline
    InternVL2-4B & Self & {\bf 67.65}   \\
    InternVL2-4B & LLaVA-1.5-7B & 65.91   \\
    \hline
    InternVL2-8B & Self & {\bf 70.10}   \\
    InternVL2-8B & LLaVA-1.5-7B & 69.49   \\
    \bottomrule
    \end{tabular}
\caption{Exchange reduction strategies of different MLLMs for Scenario~\uppercase\expandafter{\romannumeral1}.
The strategy (\ie keeping rates) of one MLLM from the search of P-Sigmoid does not generalizes to others.
}
\label{table:gen_search_results}
\end{table}

\noindent
{\bf MLLMs need flexible reduction methods.} 
For Scenario~\uppercase\expandafter{\romannumeral1}, reduction strategies should be customized for different MLLMs with the following evidences.
First, as shown in Fig.~\ref{fig:keep_rates_from_gsearch}, the S-curves of MLLMs from G-Search are different.
Since G-Search reaches the optimal solution based on our assumption (See Sec.~\ref{subsec:greedy_search}), different MLLMs have their own optimal reduction strategies.
Second, we apply the keeping rates of LLaVA-1.5-7B on InternVL2-4B and 8B models, and exchange the keeping rates of InternVL2-1B and 2B models. 
As shown in Table~\ref{table:gen_search_results}, all MLLMs suffer from performance drops when using reduction strategies (\ie keeping rates) of others.
As a result, handcrafted methods can hardly provide appropriate reduction strategies for various MLLMs, while our method is flexible and can adjust reduction strategies based on MLLMs.

\noindent
{\bf Why a steady decrease in keeping rates:}
Figure~\ref{fig:keep_rates_from_gsearch} shows that the keeping rates of G-Search decrease steadily.
This is not only because keeping rates cannot increase based on our assumption.
Moreover, as shown in Fig.~\ref{fig:attn_scres_llava_7b_small}, the median of the attention scores of vision tokens drops in deep layers of the LLM within LLaVA-1.5-7B.
Check similar visualizations for various MLLMs in the supplement.
As shown in recent studies~\citep{xiao2024efficient,li2024snapkv}, tokens of low attention scores in LLMs are not important and can be removed during inference. 
We assume that the drop in attention scores of vision tokens leads to the drop in the number of important tokens with high attention scores. 
Thus, keeping rates decrease.

\noindent
{\bf Training with reductions:}
Although the proposed G-Search and P-Sigmoid are training-free, MLLMs can be trained with reductions using keeping rates from our method. 
We collect keeping rates of a pretrained LLaVA-1.5-7B model from G-Search and P-Sigmoid for Scenario~\uppercase\expandafter{\romannumeral1} and Scenario~\uppercase\expandafter{\romannumeral2}, respectively.
Then, new LLaVA-1.5-7B models are trained or finetuned with reductions.
The budget of P-Sigmoid is set as that of FastV with R=87.5\%.
We use the same training data as LLaVA~\cite{liu2024improved} for both training and finetuning.
As shown in Table~\ref{table:train_w_reduce}, there is no performance boost in training/finetuning with reductions for Scenario~\uppercase\expandafter{\romannumeral1}. The finetuning slightly improves the performance for Scenario~\uppercase\expandafter{\romannumeral2}.
Considering the huge cost of MLLM training/finetuning, our method acts as a good plug-and-play reduction solution for MLLMs.

\begin{table}[tb]
\centering
    \begin{tabular}{lc|c}
    \toprule
    {Model} & Keeping rates & Avg. acc. $\uparrow$ \\
    \hline
    Pretrained  & \multirow{3}{*}{\shortstack{G-Search \\ for Scenario~\uppercase\expandafter{\romannumeral1}}} & {\bf 48.965}  \\
    Train w/ reduction & ~ & 48.365  \\
    Fintune w/ reduction & ~ & 48.364  \\
    \hline
    Pretrained & \multirow{3}{*}{\shortstack{P-Sigmoid \\ for Scenario~\uppercase\expandafter{\romannumeral2}}} & 46.52   \\
    Train w/ reduction & ~ & 46.40  \\
    Fintune w/ reduction  & ~ & {\bf 47.09}  \\

    \bottomrule
    \end{tabular}
\caption{Training LLaVA-1.5-7B with reductions. Keeping rates are from G-Search and P-Sigmoid applied on a pretrained LLaVA-1.5-7B model.}
\label{table:train_w_reduce}
\end{table}

%% file: sec/5_conclusion.tex
\section{Conclusion}
\label{sec:conclusion}

In this paper, we attempt to improve the computational efficiency of MLLMs in two scenarios, (\uppercase\expandafter{\romannumeral1}) Reducing computational cost without degrading the performance. 
(\uppercase\expandafter{\romannumeral2}) Improving the performance with a given budget.
We find that the relative importance of each vision token remains similar at different layers. 
Based on this finding, we come to the assumption that the number of essential top vision tokens does not increase along layers.
Accordingly, we propose two training-free solutions, \ie G-Search and P-Sigmoid, to reduce vision tokens for scenarios \uppercase\expandafter{\romannumeral1} and \uppercase\expandafter{\romannumeral2}, respectively.
Extensive experiments demonstrate that our method outperforms other vision token reduction methods in terms of both efficiency and effectiveness on top of various MLLMs and on diverse benchmarks.

%% file: sec/X_suppl.tex
\clearpage
\setcounter{page}{1}
\maketitlesupplementary

\noindent
This document supplements the main paper as follows.
\begin{itemize}
  \item Sec.~\ref{sec:g_search_detail_results} provides the detailed results of G-Search on 12 benchmarks.
  \item Sec.~\ref{sec:g_search_larger_mllm} applies our method on larger MLLMs, \eg, LLaVA-1.5-13B, and InternVL2-26B. Our method consistently boosts the efficiency of MLLMs in various sizes.
  \item Sec.~\ref{sec:g_search_tokenpacker_deco} provides the detailed results to show that our G-Search can further improve the efficiency of MLLMs on top of prompt-agnostic methods.
  \item Sec.~\ref{subsec:detailed_time_cost_reduction_method} provides more metrics for efficiency evaluations, \eg, Multiply Accumulate Operations (MACs), and per-token decoding time cost.
  \item Sec.~\ref{sbusec:comp_flashattn} compares our method with the I/O aware approach FlashAttention2~\cite{dao2022flashattention,dao2023flashattention}. Our method runs faster and is able to trade off effectiveness and efficiency.
  \item Sec.~\ref{sec:eval_video} evaluates our method on video benchmarks. The reduction strategy searched by our method on image understanding generalizes to video understanding.
  \item Sec.~\ref{sbusec:set_budget_p_sigmoid} shows how to set the budget for P-Sigmoid with theoretical analysis.
  \item Sec.~\ref{subsec:k_for_mllms} provides the values of $k$ found by our P-Sigmoid. The value varies by MLLMs.
  \item Sec.~\ref{subsec:corr_layers} illustrates the correlations of MLLM layers in terms of vision tokens sorted by attention scores. The main finding of this paper holds for various MLLMs.
  \item Sec.~\ref{subsec:limit_future_work} talks about limitations of this paper and provides potential solutions.
\end{itemize}

\begin{table*}[t]
\centering
\setlength\tabcolsep{5pt} 
\resizebox{\textwidth}{!}{
    \begin{tabular}{l|cc|ccc|ccc|ccc|ccc}
    \toprule
     \multirow{2}{*}{\shortstack{Base MLLM \\+ Method}} & ~ &  ~ & \multicolumn{3}{c|}{General VQA} & \multicolumn{3}{c|}{Knowledge} & \multicolumn{3}{c|}{OCR \& Chart} & \multicolumn{3}{c}{Vision-Centric}  \\
      ~ & \multicolumn{1}{c}{\rotatebox{90}{TFLOPs $\leftarrow$}} & \multicolumn{1}{c|}{\rotatebox{90}{Avg acc. $\rightarrow$}} & \multicolumn{1}{c}{\rotatebox{90}{MME}} & \multicolumn{1}{c}{\rotatebox{90}{MMBench}} & \multicolumn{1}{c|}{\rotatebox{90}{GQA}} & \multicolumn{1}{c}{\rotatebox{90}{MMMU}} & \multicolumn{1}{c}{\rotatebox{90}{MathVista}} & \multicolumn{1}{c|}{\rotatebox{90}{AI2D}} & \multicolumn{1}{c}{\rotatebox{90}{TextVQA}} & \multicolumn{1}{c}{\rotatebox{90}{ChartQA}} & \multicolumn{1}{c|}{\rotatebox{90}{DocVQA}} & \multicolumn{1}{c}{\rotatebox{90}{POPE}} & \multicolumn{1}{c}{\rotatebox{90}{RealWorldQA}} & \multicolumn{1}{c}{\rotatebox{90}{HallusionBench}} \\
      
    \hline
    LLaVA-1.5-7B & 9.18 & 48.97 & 1743.2 & 65.8 & 62.8 & 37.6 & 21.6 & 55.6 & 46.4 & 17.7 & 29.1 & 86.1 & 55.0 & 47.6 \\
    + VTW & \underline{5.19} & 44.32 & 1740.6 & 65.9 & 56.4 & 37.8 & 21.2 & 55.5 & 16.5 & 13.8 & 14.1 & 86.0 & 54.6 & 47.7 \\
    + PDrop & 4.95 & \underline{48.70} & 1714.9 & 65.5 & 61.9 & 38.0 & 21.9 & 55.2 & 46.4 & 17.2 & 28.9 & 86.1 & 55.5 & 46.6 \\
    + FastV (R=50\%) & 5.47 & \underline{48.70} & 1760.8 & 64.8 & 61.8 & 37.7 & 21.8 & 55.6 & 46.2 & 17.9 & 27.8 & 84.7 & 56.2 & 47.1 \\
    + G-Search (Ours) & {\bf 3.95} & {\bf 48.77} & 1741.1 & 65.3 & 62.1 & 37.7 & 22.0 & 55.5 & 46.0 & 17.8 & 28.1 & 85.5 & 56.3 & 46.7 \\
    
    \hline
    InternVL2-1B & 4.62 & 59.85 & 1778.3 & 63.2 & 55.1 & 34.6 & 32.6 & 62.6 & 69.1 & 71.3 & 82.0 & 87.3 & 51.5 & 45.3 \\
    + VTW & {\bf 3.73} & 41.13 & 1767.5 & 62.9 & 42.5 & 33.8 & 20.9 & 60.7 & 11.7 & 12.2 & 11.7 & 83.9 & 46.4 & 43.7 \\
    + PDrop & 3.88 & 53.70 & 1764.3 & 62.0 & 53.4 & 34.7 & 29.2 & 59.5 & 59.3 & 51.0 & 52.4 & 86.7 & 49.9 & 43.3 \\
    + FastV (R=50\%) & 3.91 & \underline{54.85} & 1747.4 & 61.8 & 53.8 & 34.9 & 30.9 & 59.8 & 62.1 & 58.7 & 55.8 & 85.1 & 48.4 & 44.6 \\
    + G-Search (Ours) & \underline{3.84} & {\bf 59.19} & 1750.7 & 62.5 & 55.0 & 34.8 & 32.5 & 62.1 & 68.3 & 69.4 & 79.0 & 87.0 & 51.9 & 45.2 \\
    
    \hline
    InternVL2-2B & 8.10 & 61.94 & 1821.7 & 72.5 & 59.9 & 34.7 & 33.8 & 72.5 & 72.0 & 75.0 & 87.2 & 88.4 & 34.0 & 48.3 \\
    + VTW & {\bf 5.50} & 37.84 & 1596.3 & 64.2 & 42.6 & 32.3 & 25.0 & 66.8 & 10.7 & 11.0 & 11.3 & 67.8 & 22.9 & 42.5 \\
    + PDrop & 5.93 & 58.98 & 1800.1 & 71.5 & 58.2 & 33.9 & 32.6 & 70.6 & 70.2 & 67.6 & 73.7 & 85.9 & 33.2 & 46.2 \\
    + FastV (R=50\%) & 5.85 & \underline{59.91} & 1774.2 & 71.8 & 58.7 & 34.1 & 33.1 & 71.6 & 70.7 & 69.6 & 75.7 & 88.0 & 32.9 & 49.3 \\
    + G-Search (Ours) & \underline{5.64} & {\bf 61.22} & 1831.7 & 71.9 & 59.4 & 34.8 & 33.6 & 71.9 & 71.6 & 71.6 & 84.4 & 88.2 & 34.0 & 47.7 \\
    
    \hline
    InternVL2-4B & 13.97 & 68.16 & 2084.4 & 77.6 & 62.0 & 45.8 & 36.4 & 77.8 & 74.1 & 81.0 & 89.6 & 87.1 & 59.7 & 52.3 \\
    + VTW & 8.72 & 49.01 & 2027.9 & 76.4 & 51.1 & 45.8 & 26.7 & 77.3 & 14.7 & 14.8 & 15.2 & 85.2 & 56.5 & 52.0 \\
    + PDrop & {\bf 8.35} & \underline{66.94} & 2084.5 & 77.4 & 61.4 & 46.3 & 35.5 & 77.0 & 72.7 & 77.4 & 82.2 & 87.0 & 60.9 & 50.9 \\
    + FastV (R=50\%) & 9.01 & 66.19 & 2080.1 & 77.6 & 61.6 & 45.9 & 35.2 & 77.5 & 72.5 & 75.1 & 76.5 & 86.7 & 60.0 & 51.5 \\
    + G-Search (Ours) & \underline{8.69} & {\bf 67.65} & 2082.3 & 77.4 & 61.8 & 46.0 & 36.4 & 77.5 & 73.9 & 80.2 & 86.9 & 87.0 & 59.5 & 50.8 \\
    
    \hline
    InternVL2-8B & 24.10 & 70.83 & 2205.3 & 81.8 & 62.7 & 48.6 & 36.9 & 82.4 & 76.6 & 82.6 & 91.9 & 86.7 & 65.5 & 55.5 \\
    + VTW & 13.71 & 52.51 & 2195.1 & 81.6 & 56.6 & 46.7 & 31.4 & 81.8 & 18.9 & 16.3 & 17.5 & 85.8 & 63.4 & 51.7 \\
    + PDrop & \underline{13.13} & 69.19 & 2193.1 & 81.4 & 62.3 & 45.8 & 35.7 & 80.4 & 75.6 & 81.6 & 86.3 & 86.7 & 64.6 & 51.5 \\
    + FastV (R=50\%) & 14.58 & \underline{69.42} & 2214.2 & 81.2 & 62.0 & 48.1 & 37.3 & 81.1 & 75.6 & 80.2 & 83.4 & 86.5 & 64.6 & 53.8 \\
    + G-Search (Ours) & {\bf 12.24} & {\bf 70.10} & 2216.7 & 81.4 & 62.6 & 48.6 & 36.2 & 82.1 & 76.0 & 81.1 & 89.8 & 86.9 & 65.1 & 52.3 \\
    
    \hline
    \hline
    LLaVA-1.5-13B & 17.44 & 49.77 & 1818.0 & 68.8 & 63.3 & 35.6 & 22.9 & 59.3 & 47.6 & 18.2 & 30.3 & 85.9 & 55.0 & 45.4 \\
    + VTW & 9.77 & 46.93 & 1828.1 & 68.7 & 60.1 & 35.4 & 22.6 & 59.3 & 33.7 & 15.5 & 15.1 & 86.0 & 55.7 & 45.6 \\
    + PDrop & \underline{9.29} & 49.24 & 1810.4 & 68.4 & 63.0 & 36.1 & 23.3 & 59.1 & 47.6 & 18.3 & 23.5 & 86.0 & 55.3 & 45.6 \\
    + FastV (R=50\%) & 10.18 & {\bf 49.69} & 1857.4 & 68.4 & 62.6 & 36.1 & 23.5 & 58.9 & 47.2 & 18.3 & 28.9 & 85.0 & 55.7 & 45.4 \\
    + G-Search (Ours) & {\bf 6.45} & \underline{49.65} & 1835.5 & 68.2 & 62.4 & 36.6 & 23.7 & 58.9 & 47.6 & 18.5 & 29.0 & 85.8 & 54.4 & 45.1 \\

    \hline
    InternVL2-26B & 111.45 & 74.12 & 2272.8 & 81.8 & 65.2 & - & 37.8 & 83.1 & 82.0 & 84.7 & 90.5 & 88.0 & 67.7 & 53.4 \\
    + VTW & 84.94 & 67.12 & 2293.4 & 81.8 & 63.3 & - & 38.0 & 83.3 & 55.3 & 61.6 & 63.6 & 88.0 & 66.7 & 54.7 \\
    + PDrop & \underline{83.30} & \underline{73.12} & 2245.8 & 81.6 & 65.2 & - & 37.2 & 81.9 & 81.8 & 84.0 & 83.2 & 88.2 & 67.5 & 53.6 \\
    + FastV (R=50\%) & 86.25 & 72.75 & 2244.6 & 81.3 & 64.8 & - & 37.5 & 82.1 & 81.0 & 83.0 & 82.5 & 87.4 & 67.2 & 53.4 \\
    + G-Search (Ours) & {\bf 78.98} & {\bf 73.72} & 2253.5 & 81.7 & 65.1 & - & 37.7 & 82.6 & 81.8 & 84.4 & 88.0 & 88.2 & 67.2 & 53.8 \\
    
    \bottomrule
    \end{tabular}
}
\caption{Detailed results of prompt-aware methods on 12 benchmarks. InternVL2-26B on MMMU is not reported due to the out-of-memory issue on our platform. Average accuracy for InternVL2-26B is averaged on the other 11 benchmarks}
\label{table:comp_prompt_aware_details}
\end{table*}

\begin{table*}[t]
\centering
\setlength\tabcolsep{6pt} 
\resizebox{\textwidth}{!}{
    \begin{tabular}{l|rr|rrr|rrr|rrr|rrr}
    \toprule
     \multirow{2}{*}{\shortstack{LLaVA-1.5-7B\\+ Method \ \ \ \ \ \ \ }}  & \multicolumn{1}{c}{} &  \multicolumn{1}{c|}{} & \multicolumn{3}{c|}{General VQA} & \multicolumn{3}{c|}{Knowledge} & \multicolumn{3}{c|}{OCR \& Chart} & \multicolumn{3}{c}{Vision-Centric}  \\
       & \multicolumn{1}{c}{\rotatebox{90}{TFLOPs $\leftarrow$}} & \multicolumn{1}{c|}{\rotatebox{90}{Avg acc. $\rightarrow$}} & \multicolumn{1}{c}{\rotatebox{90}{MME}} & \multicolumn{1}{c}{\rotatebox{90}{MMBench}} & \multicolumn{1}{c|}{\rotatebox{90}{GQA}} & \multicolumn{1}{c}{\rotatebox{90}{MMMU}} & \multicolumn{1}{c}{\rotatebox{90}{MathVista}} & \multicolumn{1}{c|}{\rotatebox{90}{AI2D}} & \multicolumn{1}{c}{\rotatebox{90}{TextVQA}} & \multicolumn{1}{c}{\rotatebox{90}{ChartQA}} & \multicolumn{1}{c|}{\rotatebox{90}{DocVQA}} & \multicolumn{1}{c}{\rotatebox{90}{POPE}} &  \multicolumn{1}{c}{\rotatebox{90}{RealWorldQA}} & \multicolumn{1}{c}{\rotatebox{90}{HallusionBench}} \\
       
    \hline
    TokenPacker & 3.27 & 47.60 & 1726.1 & 64.5 & 61.7 & 37.1 & 21.9 & 55.0 & 41.7 & 16.8 & 23.2 & 86.1 & 53.5 & 48.0 \\
    + G-Search (Ours) & 2.12 & 47.68 & 1756.4 & 64.6 & 61.3 & 37.1 & 22.0 & 55.5 & 41.8 & 16.2 & 22.8 & 86.8 & 52.5 & 48.8 \\
    \multicolumn{1}{c}{$\Delta$} & {\bf -1.15} & {\bf+0.08} & {\bf +30.3} & {\bf +0.1} & -0.4 & 0.0 & {\bf +0.1} & {\bf +0.5} & {\bf +0.1} & -0.6 & -0.4 & {\bf +0.7} & -1.0 & {\bf +0.8} \\

    \hline
    DeCo & 3.26 & 46.97 & 1714.1 & 64.3 & 61.4 & 37.2 & 21.4 & 54.8 & 40.1 & 15.8 & 22.4 & 84.9 & 53.3 & 46.9 \\
    + G-Search (Ours) & 2.16 & 46.71 & 1697.6 & 64.5 & 60.7 & 37.0 & 21.1 & 54.7 & 40.0 & 15.6 & 22.4 & 85.0 & 52.4 & 46.5 \\
    \multicolumn{1}{c}{$\Delta$} & {\bf -1.10} & -0.26 & -16.5 & {\bf +0.2} & -0.7 & -0.2 & -0.3 & -0.1 & -0.1 & -0.2 & 0.0 & {\bf +0.1} & -0.9 & -0.4 \\

    \bottomrule
    \end{tabular}
}
\caption{Detailed results for G-Search applied on top of prompt-agnostic methods.}
\label{table:improve_prompt_agnostic}
\end{table*}

\section{Detailed results on 12 benchmarks}

\subsection{G-Search and existing prompt-aware methods}\label{sec:g_search_detail_results}
Table~\ref{table:comp_prompt_aware_details} provides the concrete results on the 12 image understanding benchmarks for the proposed G-Search and several prompt-aware vision token reduction methods, \ie, VTW~\citep{lin2024boosting}, PDrop~\cite{xing2024pyramiddrop}, and FastV~\citep{chen2024image}, on top of various MLLMs.
The results are complementary to Table~\ref{table:run_time_eval_s1} of the main paper.
Our method consistently reduce computational cost without significant performance drops, while other methods may fail in preserving the performance.

\subsection{G-Search on larger MLLMs.}\label{sec:g_search_larger_mllm}
Besides MLLMs explored in the main paper, we apply G-Search on top of larger MLLMs, \eg, LLaVA-1.5-13B and InternVL2-26B, and report the results in the last two blocks of Table~\ref{table:comp_prompt_aware_details}.
The performance of InternVL2-26B on MMMU is not reported because it runs out of GPU memories during inference probably due to too long contexts. Thus, we compute the average accuracy for InternVL2-26B on the rest 11 benchmarks.
The results on larger MLLMs are in line with those on smaller MLLMs.
That is, the proposed G-Search requires less computations without a significant performance drop, while other methods either require much more computations or suffer from performance drops.
Such results indicate that our method is robust on various sizes of models and scales up well.

\subsection{G-Search on top of prompt-agnostic methods}\label{sec:g_search_tokenpacker_deco}
Table~\ref{table:improve_prompt_agnostic} provides the concrete results of applying the proposed G-Search on two prompt-agnostic vision token reduction methods, \ie, TokenPacker~\citep{li2024tokenpacker}, and DeCo~\citep{yao2024deco}.
The results are complementary to Table~\ref{table:run_time_eval_on_prompt_agnostic} of the main paper.
As one can see, our G-Search significantly reduces the computational cost with slight variances in performance on all benchmarks, which clearly demonstrates that our method is flexible and robust in various cases.

\begin{table*}[tb]
\centering
\setlength\tabcolsep{4.5pt} 
    \begin{tabular}{l|cc|ccc|cc}
    \toprule
    \multirow{2}{*}{\shortstack{Base MLLM \\+ Method}} & Average & Memory  & \multirow{2}{*}{TFLOPs$\downarrow$} & \multirow{2}{*}{TMACs$\downarrow$} & \multirow{2}{*}{\# Params (B)} & Prefilling  & Decoding  \\
    ~ & accuracy$\uparrow$ & cost$\downarrow$ & ~ & ~ & ~ & time cost$\downarrow$ & time cost$\downarrow$ \\
    \hline
    
    LLaVA-1.5-7B & 48.97 & 1.000 & 9.18 & 4.59 & 6.76 & 0.625 & 0.181  \\
    + VTW & 44.32 & 0.500 & 5.19 & 2.60 & 6.76 & 0.385 & 0.134  \\
    + PDrop & 48.70 & 0.469 & 4.95 & 2.47 & 6.76 & 0.381 & 0.123  \\
    + FastV (R=50\%) & 48.70 & 0.531 & 5.47 & 2.73 & 6.76 & 0.387 & 0.136  \\
    + G-Search (Ours) & 48.77 & 0.340 & 3.95 & 1.98 & 6.76 & 0.301 & 0.117  \\
    
    \hline
    InternVL2-1B & 59.85 & 1.000 & 4.62 & 2.31 & 0.94 & 0.384 & 0.123  \\
    + VTW & 41.13 & 0.500 & 3.73 & 1.86 & 0.94 & 0.331 & 0.114  \\
    + PDrop & 53.70 & 0.583 & 3.88 & 1.94 & 0.94 & 0.336 & 0.114  \\
    + FastV (R=50\%) & 54.85 & 0.542 & 3.91 & 1.96 & 0.94 & 0.342 & 0.120  \\
    + G-Search (Ours) & 59.19 & 0.527 & 3.84 & 1.92 & 0.94 & 0.333 & 0.114  \\
    
    \hline
    InternVL2-2B & 61.94 & 1.000 & 8.10 & 4.05 & 2.21 & 0.598 & 0.172  \\
    + VTW & 37.84 & 0.500 & 5.50 & 2.75 & 2.21 & 0.439 & 0.138  \\
    + PDrop & 58.98 & 0.583 & 5.93 & 2.96 & 2.21 & 0.452 & 0.140  \\
    + FastV (R=50\%) & 59.91 & 0.542 & 5.85 & 2.92 & 2.21 & 0.451 & 0.135  \\
    + G-Search (Ours) & 61.22 & 0.532 & 5.64 & 2.82 & 2.21 & 0.444 & 0.132  \\
    
    \hline
    InternVL2-4B & 68.16 & 1.000 & 13.97 & 6.98 & 4.15 & 0.969 & 0.256  \\
    + VTW & 49.01 & 0.500 & 8.72 & 4.36 & 4.15 & 0.649 & 0.189  \\
    + PDrop & 66.94 & 0.469 & 8.35 & 4.17 & 4.15 & 0.627 & 0.190  \\
    + FastV (R=50\%) & 66.19 & 0.531 & 9.01 & 4.50 & 4.15 & 0.652 & 0.199  \\
    + G-Search (Ours) & 67.65 & 0.488 & 8.69 & 4.34 & 4.15 & 0.645 & 0.190  \\
    
    \hline
    InternVL2-8B & 70.83 & 1.000 & 24.10 & 12.05 & 8.08 & 1.518 & 0.388  \\
    + VTW & 52.51 & 0.500 & 13.71 & 6.85 & 8.08 & 0.927 & 0.251  \\
    + PDrop & 69.19 & 0.469 & 13.13 & 6.56 & 8.08 & 0.915 & 0.249  \\
    + FastV (R=50\%) & 69.42 & 0.531 & 14.58 & 7.29 & 8.08 & 0.998 & 0.266  \\
    + G-Search (Ours) & 70.10 & 0.424 & 12.24 & 6.12 & 8.08 & 0.860 & 0.237  \\
    
    \bottomrule
    \end{tabular}
\caption{
More metrics of efficiency for Scenario~\uppercase\expandafter{\romannumeral1}. 
We report the average accuracy calculated on 12 benchmarks and the per-token decoding time cost. Our method achieves good decoding time cost, as well as prefilling time cost.
}
\label{table:run_time_eval_s1_detail}
\end{table*}

\begin{table*}[tb]
\centering
\setlength\tabcolsep{4.5pt} 
    \begin{tabular}{l|cc|ccc|cc}
    \toprule
    \multirow{2}{*}{\shortstack{Base MLLM \\+ Method}} & Average & Memory  & \multirow{2}{*}{TFLOPs$\downarrow$} & \multirow{2}{*}{TMACs$\downarrow$} & \multirow{2}{*}{\# Params (B)} & Prefilling  & Decoding  \\
    ~ & accuracy$\uparrow$ & cost$\downarrow$ & ~ & ~ & ~ & time cost$\downarrow$ & time cost$\downarrow$ \\
    \hline
    LLaVA-1.5-7B & 48.97 & 1.000 & 9.18 & 4.59 & 6.76 & 0.625 & 0.181  \\
    + FastV  & 43.14 & 0.180 & 2.74 & 1.37 & 6.76 & 0.227 & 0.098  \\
    + P-Sigmoid (Ours) & 46.52 & 0.171 & 2.66 & 1.33 & 6.76 & 0.223 & 0.095  \\
    \hline
    InternVL2-1B & 59.85 & 1.000 & 4.62 & 2.31 & 0.94 & 0.384 & 0.123  \\
    + FastV & 43.16 & 0.198 & 3.43 & 1.71 & 0.94 & 0.308 & 0.108  \\
    + P-Sigmoid (Ours) & 47.83 & 0.188 & 3.38 & 1.69 & 0.94 & 0.304 & 0.106  \\
    \hline
    InternVL2-2B & 61.94 & 1.000 & 8.10 & 4.05 & 2.21 & 0.598 & 0.172  \\
    + FastV & 47.66 & 0.198 & 4.26 & 2.13 & 2.21 & 0.353 & 0.112  \\
    + P-Sigmoid (Ours) & 51.54 & 0.188 & 4.16 & 2.08 & 2.21 & 0.350 & 0.109  \\
    \hline
    InternVL2-4B & 68.16 & 1.000 & 13.97 & 6.98 & 4.15 & 0.969 & 0.256  \\
    + FastV & 54.83 & 0.180 & 5.48 & 2.74 & 4.15 & 0.441 & 0.143  \\
    + P-Sigmoid (Ours) & 61.19 & 0.171 & 5.38 & 2.69 & 4.15 & 0.436 & 0.141  \\
    \hline
    InternVL2-8B & 70.83 & 1.000 & 24.10 & 12.05 & 8.08 & 1.518 & 0.388  \\
    + FastV & 55.17 & 0.180 & 7.71 & 3.86 & 8.08 & 0.590 & 0.182  \\
    + P-Sigmoid (Ours) & 62.86 & 0.171 & 7.46 & 3.73 & 8.08 & 0.577 & 0.178  \\
    
    \bottomrule
    \end{tabular}
\caption{More metrics of efficiency for  Scenario~\uppercase\expandafter{\romannumeral2}. We set R=87.5\% for Fast V.}
\label{table:run_time_eval_s2}
\end{table*}

\begin{table*}[tb]
\centering
\setlength\tabcolsep{4.5pt} 
    \begin{tabular}{l|c|ccc|cc}
    \toprule
    \multirow{2}{*}{\shortstack{Base MLLM \\+ Method}} & Average  & \multirow{2}{*}{TFLOPs$\downarrow$} & \multirow{2}{*}{TMACs$\downarrow$} & \multirow{2}{*}{\# Params (B)} & Prefilling  & Decoding  \\
    ~ & accuracy$\uparrow$ & ~ & ~ & ~ & time cost$\downarrow$ & time cost$\downarrow$ \\
    \hline
    LLaVA-1.5-7B & 48.43  & 9.18 & 4.59 & 6.76 & 0.135 & 0.077  \\
    + FlashAttention2 & 48.41  & 8.96 & 4.48 & 6.76 & 0.128 & 0.073  \\
    + G-Search (Ours) & 48.25  & 3.95 & 1.98 & 6.76 & 0.109 & 0.068  \\
    \hline
    InternVL2-8B & 70.39 & 24.10 & 12.05 & 8.08 & 0.278 & 0.101  \\
    + FlashAttention2 & 70.34 & 23.67 & 11.84 & 8.08 & 0.212 & 0.085  \\
    + G-Search (Ours) & 70.07 & 12.24 & 6.12  & 8.08 & 0.163 & 0.074  \\
    
    \bottomrule
    \end{tabular}
\caption{Comparison to FlashAttention2. Bflot16 is adopted to enable FlashAttention2. Our method achieves better efficiency with negligible performance drops. Furthermore, our method is able to trade off the efficiency and the performance for Scenario~\uppercase\expandafter{\romannumeral2}.}
\label{table:runtime_flashattn}
\end{table*}

\section{Extra evaluations of computational costs}\label{sec:detaled_time_cost}

Not only can our methods reduce the number of vision tokens to accelerate the prefilling phrase of the inference, but also it reduces the KV-Cache to speed up the decoding phrase.
In addition to the prefilling time cost in the main paper, we report more metrics for computational costs in this section, including MACs, and the per-token time cost at the decoding stage.

\subsection{Comparison to existing reduction methods.} \label{subsec:detailed_time_cost_reduction_method}
Table~\ref{table:run_time_eval_s1_detail} provides more metrics to evaluate the efficiency for Scenario~\uppercase\expandafter{\romannumeral1}.
The proposed G-Search is compared with other completing reduction methods, \ie, VTW~\citep{lin2024boosting}, PDrop~\cite{xing2024pyramiddrop}, and FastV~\citep{chen2024image}.
The results are complementary to Table~\ref{table:run_time_eval_s1} of the main paper.
As shown in Table~\ref{table:run_time_eval_s1_detail}, our method achieves good per-token decoding time cost, as well as other metrics like prefilling time cost.

Table~\ref{table:run_time_eval_s2} provides more efficiency metrics for Scenario~\uppercase\expandafter{\romannumeral2}.
As a complement to Table~\ref{table:p_sigmoid_diff_mllm} of the main paper, we compare P-Sigmoid with the prior SOTA FastV (R=87.5\%) on top of various MLLMs.
The budget of P-Sigmoid is set similar as FastV.
As we can see, compared to FastV, P-Sigmoid achieves much better performance with similar or slightly less memory cost, TFLOPs, MACs and time costs.

\subsection{Comparison to FlashAttention.} \label{sbusec:comp_flashattn}
FlashAttention~\cite{dao2022flashattention} is a widely adopted I/O aware approach to accelerate transformer-based models like LLMs.
Unlike token reduction methods, it lowers down the number of times to read and write memories instead of reducing FLOPs.
We compare our method to the latest version of FlashAttention, \ie FlashAttention2~\cite{dao2023flashattention}, on top of LLaVA-1.5-7B and InternVL2-8B.
We evaluate models with half-precision floating-point (specifically bfloat16 is used) because FlashAttention2 only supports this data format.
Thus, the time cost of MLLMs in this section is lower than that in the main paper.
As shown in Table~\ref{table:runtime_flashattn}, FlashAttention2 has slightly lower TFLOPs than the vanilla model. This is probably because FlashAttention2 dose not require the 4D attention masks to enable casual attentions.
Although FlashAttention2 reduces time costs in both prefilling and decoding stages compared to the vanilla model, our method is faster without significant performance drops.
Moreover, our method can be further enhanced with FlashAttention2 in the decoding stage.
Besides, FlashAttention2 can only improve the efficiency for Scenario~\uppercase\expandafter{\romannumeral1} where models are accelerated without performance drops.
In contrast, our method works for both Scenario~\uppercase\expandafter{\romannumeral1} and Scenario~\uppercase\expandafter{\romannumeral2}.
We believe Scenario~\uppercase\expandafter{\romannumeral2}, where the performance is improved with given budgets, is in demand and important for edge applications, but it is ignored by current studies.

\begin{table*}[tb]
\centering
    \begin{tabular}{lc|cc|ccc}
    \toprule
     \multirow{2}{*}{\shortstack{Base MLLM \\+ Method}} & \multirow{2}{*}{\shortstack{Reduction \\ Strategy}} & \multicolumn{2}{c|}{Effectiveness} & \multicolumn{3}{c}{Efficiency} \\
    \cline{3-7}
    ~ & ~ & MVBench$\uparrow$ & Video-MME$\uparrow$ & Memory cost$\downarrow$ & TFLOPs$\downarrow$ & Time cost$\downarrow$ \\

    \hline
    InternVL2-8B & None & 64.67 & 52.41 & 1.0 & 38.71 &  2.522  \\
    \hline
    
    \hline
    + FastV (R=50\%) & Handcrafted & 64.72 & 52.48 & 0.531 & 22.91 &  1.512  \\
    + G-Search (Ours) & Automatic & 64.65 & 52.52 & 0.424 & 19.13 &  1.340  \\
    \multicolumn{1}{c}{$\Delta$} & - & -0.07 & {\bf +0.04} & {\bf -0.107} & {\bf -3.78} &  {\bf -0.172}  \\

    \hline
    + FastV (R=87.5\%) & Handcrafted & 63.88 & 51.63 & 0.180 & 11.74 & 0.891 \\
    + P-Sigmoid (Ours) & Automatic & 64.70 & 52.48 & 0.171 & 11.34 & 0.886 \\
    \multicolumn{1}{c}{$\Delta$} & - & {\bf +0.82} & {\bf +0.85} & {\bf -0.009} & {\bf -0.40} &  {\bf -0.005}  \\
    
    \bottomrule
    \end{tabular}
\caption{Evaluation two video benchmarks. 
Reduction methods are applied on top of InternVL2-8B that is trained with video data.
G-Search achieves better efficiency compared to FastV with the default setting.
P-Sigmod uses a similar budget as FastV (R=87.5\%) and gains better performance.
}
\label{table:internvl2_video}
\end{table*}

\section{Evaluation on video benchmarks}\label{sec:eval_video}
\noindent
It is a common practice that MLLMs handle a video by sampling several images from the video and encoding each image into vision tokens.
Such an approach leads to a large amount of redundant vision tokens, and should benefit from vision token reduction methods.
In this section, we demonstrate our reduction method also accelerates MLLMs in video understanding.
We evaluate our method on top of InternVL2-8B, which is trained with videos, on two popular video understanding benchmarks, \ie, MVBench~\cite{li2024mvbench} and Video-MME~\cite{fu2024video}.
The InternVL2 model will sample 8 images from the input video and encode them into more than two thousands of vision tokens.
We leverage the reduction strategy from our G-Search, which finds the optimal keeping rates on image understanding data, to speed up the InternVL2 model.

Table~\ref{table:internvl2_video} compares our method to FastV variants configured as R=50\% and R=87.5\%. 
We have the following interesting findings.
First, we can reduce more computational costs for video understanding tasks than image understanding tasks. 
For example, P-Sigmoid requires less TFLOPs and runs faster than G-search. But they gain similar performance on the two video benchmarks.
This is plausible because there more redundant vision tokens in videos than in images. Thus, more tokens can be removed without information loss.
Second, for Scenario~\uppercase\expandafter{\romannumeral2}, 
the performance gap between P-Sigmoid and FastV in video understanding is smaller than that in image understanding.
Probably, this is caused by the fact that videos gain lots of redundant vision tokens. Non-optimal reduction strategy is likely to remove tokens without much information loss.
Such results motivate a promising future work that explores how to remove highly redundant vision tokens.

\section{More analysis \& illustration}\label{sec:extra_visualization}

\subsection{How to set budgets for P-Sigmoid} \label{sbusec:set_budget_p_sigmoid}
As mentioned in the main paper, to set the budgets of P-Sigmoid as the budgets of FastV~\cite{chen2024image}, we first set the number of vision tokens in P-Sigmoid the same as that in FastV. Then, we slightly lower down the number of vision tokens to match the TFLOPs of P-Sigmoid and FastV.
This is because the number of vision tokens is not exactly proportional to TFLOPs, as discussed below.

Per the discussion in FastV~\cite{chen2024image}, the total FLOPs of $i$-th layer of a LLM is $C_i = 4n_id^2 + 2n_i^2d + 2n_idm$ where $n_i$ is the number of tokens at this layer, $d$ is the hidden state size of the multi-head attention, and $m$ is the intermediate size of the feed-forward network.
Thus, for a LLM with $L$ layers, the total FLOPs $C$ can be written as,
\begin{align} \label{eq:target_f}
    C &= \sum_i^L(4n_id^2 + 2n_i^2d + 2n_idm) \\
    &= (4d^2 + 2dm)N +\sum_i^L n_i^2
\end{align}
where $N = \sum_i n_i$ refers to the number of total tokens from all layers.
The term $\sum_i^L n_i^2$ reaches the minimal when $\forall n_i = N/L$, which is the case of FastV. 
We provide a brief proof for the above statement below.
According to Cauchy–Schwarz inequality,
\begin{align} \label{eq:target_f}
    (\sum_i^n x_i y_i)^2 \leq (\sum_i^n x_i^2)(\sum_i^n y_i^2) .
\end{align}
When the inequality becomes an equality if and only if $\forall i,\forall j, x_i/y_i = x_j/y_j$.
We set $y_i = 1$, $x_i=n_i$ and $n=L$, and we have 
\begin{align} \label{eq:target_f}
    L(\sum_i^L n_i^2) \geq (\sum_i^L n_i)^2 = N^2 .
\end{align}
When $\forall n_i = N/L$, $(\sum_i^L n_i^2)$ reaches the minimal $ N^2/L$.

Since FastV reaches the minimal FLOPs for a given $N$, any other reduction strategies always have more FLOPs for the same $N$.
Therefore, to match the FLOPs of FastV and our method, we have to reduce the number of total tokens $N$ for our method.

\begin{figure}[t]
    \centering
    \includegraphics[width=1.0\columnwidth]{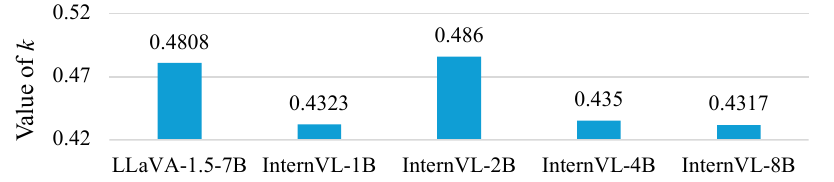}
    \caption{Values of $k$ for different MLLMs from P-Sigmoid.}
    \label{fig:k_vs_mllm_given_budget}
\end{figure}

\subsection{Values of $k$ for different MLLMs} \label{subsec:k_for_mllms}
Fig.~\ref{fig:k_vs_mllm_given_budget} illustrates values of $k$ from P-Sigmoid for Table~\ref{table:run_time_eval_s2}. 
As shown, on top of LLaVA-1.5-7B and InternVL-8B, our method has the same memory cost but outputs different $k$.
On top of InternVL-1B and InternVL-2B, P-Sigmoid again outputs different $k$ with the same memory cost.
Those results indicate that different MLLMs should have different parameters and reduction strategies, which further explains why P-Sigmoid with automatic search can outperform other methods regardless of MLLMs.

\begin{figure*}[t]
\centering
     \begin{subfigure}[b]{1.0\textwidth}
         \centering
         \includegraphics[width=0.47\textwidth]{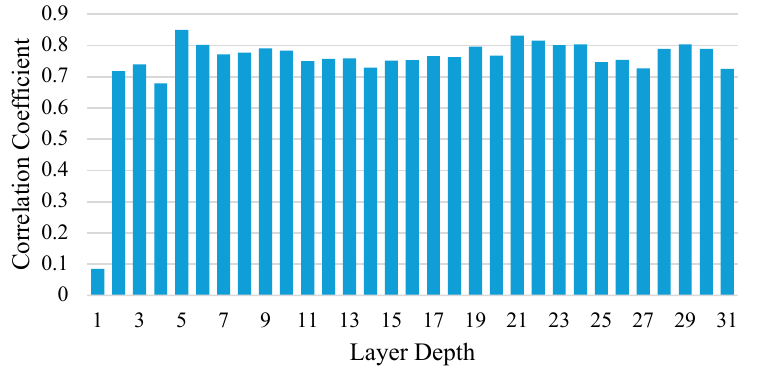}
         \caption{
         LLaVA-1.5-7B
         }
         \label{fig:supp_attn_corr_llava_7b}
     \end{subfigure}
     \begin{subfigure}[b]{1.0\textwidth}
         \centering
         \includegraphics[width=0.47\textwidth]{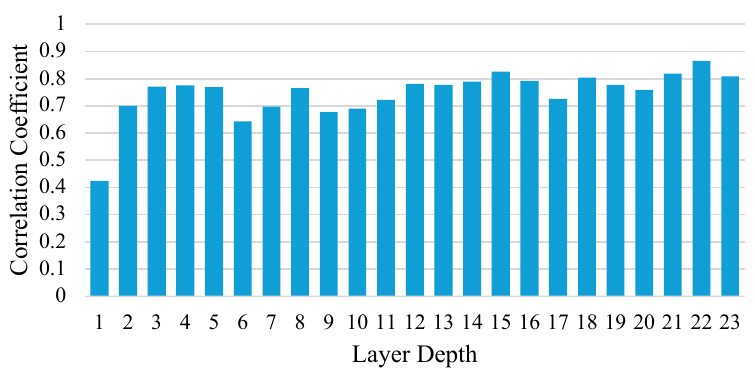}
         \hfill
         \includegraphics[width=0.47\textwidth]{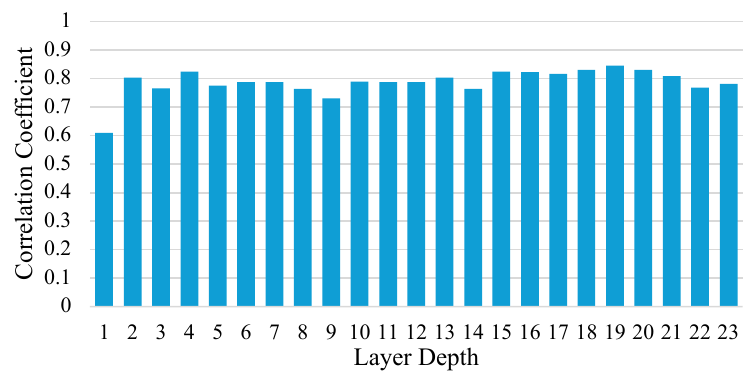}
         \hfill
         \includegraphics[width=0.47\textwidth]{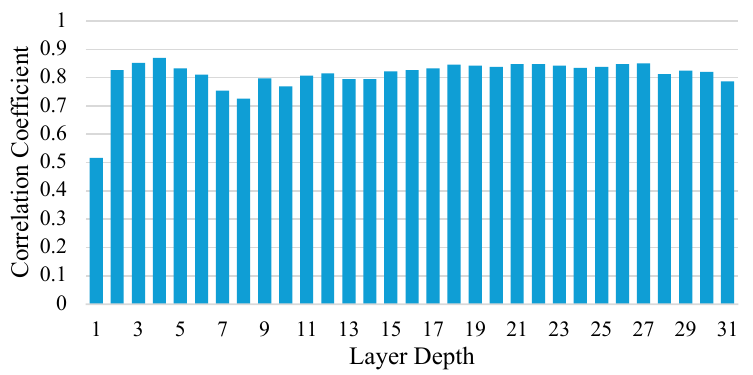}
         \hfill
         \includegraphics[width=0.47\textwidth]{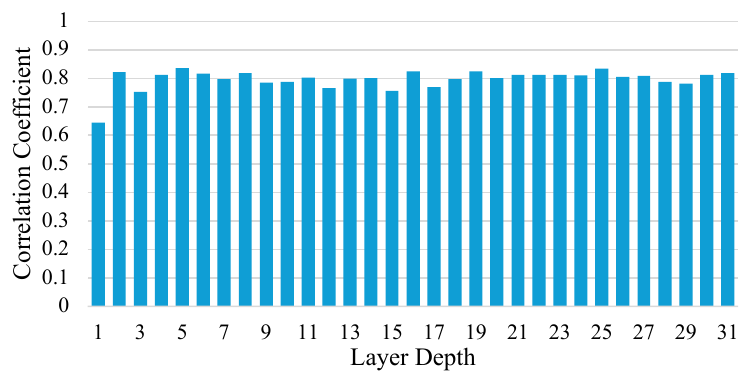}
         \caption{
         InternVL2-1B (top left), 2B (top right), 4B (bottom left), 8B (bottom right)
         }
         \label{fig:supp_attn_corr_internvl2}
     \end{subfigure}
    \caption{
     {\bf (a)}: Kendall’s Tau correlation coefficient between the current layer and the next layer of LLaVA-1.5-7B
     {\bf (b)}: Kendall’s Tau correlation coefficients for InternVL2 family. Our finding on LLaVA-1.5-7B hold on InternVL2 family.
     }
     \label{fig:supp_attn_corr_diff_mllm}
\end{figure*}

\begin{figure*}[t]
\centering
    \begin{subfigure}[b]{0.43\textwidth}
        \includegraphics[width=\textwidth]{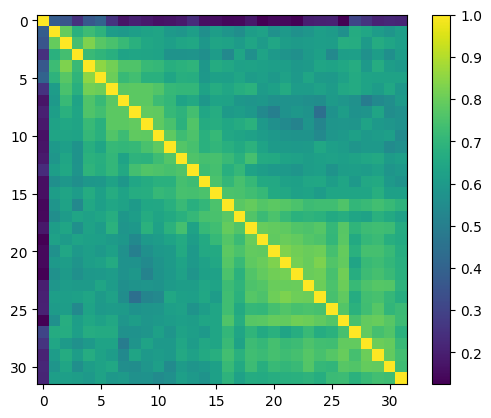} 
         \caption{
         LLaVA-1.5-7B
         }
         \label{fig:supp_corr_map_llava_7b}
    \end{subfigure}
    \hfill
    \\
    \begin{subfigure}[b]{0.43\textwidth}
        \includegraphics[width=\textwidth]{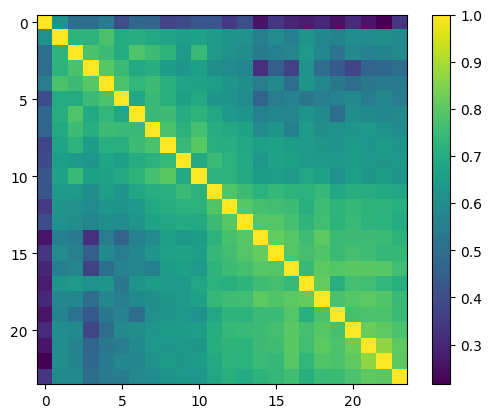} 
         \caption{
         InternVL2-1B
         }
         \label{fig:supp_corr_map_internvl2_1b}
    \end{subfigure}
    \hfill
    \begin{subfigure}[b]{0.43\textwidth}
        \includegraphics[width=\textwidth]{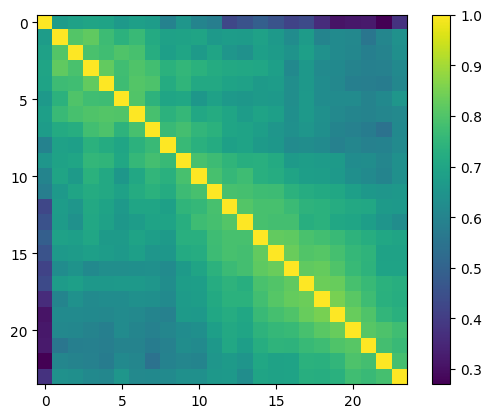}
         \caption{
         InternVL2-2B
         }
         \label{fig:supp_corr_map_internvl2_2b}
    \end{subfigure}
    \\
    \begin{subfigure}[b]{0.43\textwidth}
        \includegraphics[width=\textwidth]{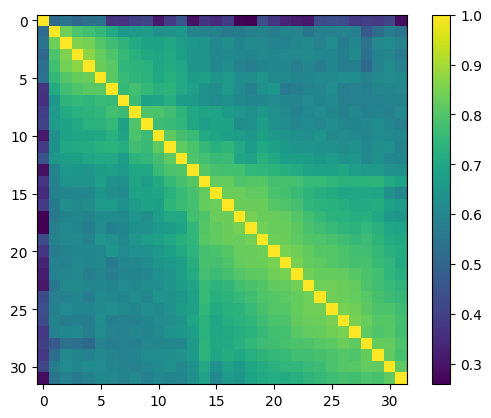}
         \caption{
         InternVL2-4B
         }
         \label{fig:supp_corr_map_internvl2_4b}
    \end{subfigure}
    \hfill
    \begin{subfigure}[b]{0.43\textwidth}
        \includegraphics[width=\textwidth]{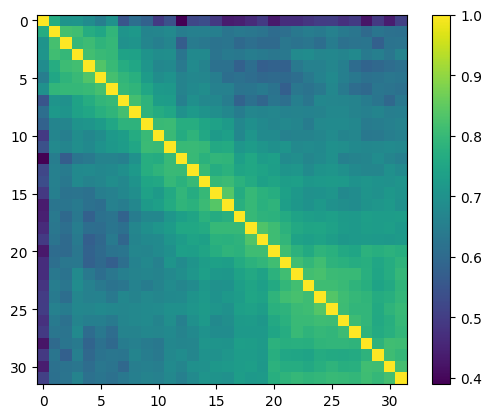}
         \caption{
         InternVL2-8B
         }
         \label{fig:supp_corr_map_internvl2_8b}
    \end{subfigure}
    \caption{
     Kendall’s Tau correlation coefficient between every two layers of various MLLMs.
     }
     \label{fig:supp_corr_map_diff_mllm}
\end{figure*}

\subsection{Correlations between layers} \label{subsec:corr_layers}
In Sec.~\ref{sec:intro} of the main paper, we demonstrate our main finding by analyzing LLaVA-1.5-7B.
As shown in Fig.~\ref{fig:supp_attn_corr_llava_7b}, we sort the vision tokens of each layer of LLaVA-1.5-7B based on their attention scores to instruction tokens and calculate the Kendall's Tau correlation coefficient~\cite{kendall1938new} between the current layer and the next layer.
We regard the relative importance of one vision token as its ranking.
Then, the finding is that the relative importance of each vision token remains similar in each layer of MLLMs after the first layer.
In this section, we provide the same visualization by analyzing InternVL2-1B/2B/4B/8B models.
As shown in Fig.~\ref{fig:supp_attn_corr_internvl2}, we get similar observations as Fig.~\ref{fig:supp_attn_corr_llava_7b}.
As a result, our finding is a general case for different MLLMs.

In addition to correlation between consecutive layers, Fig.~\ref{fig:supp_corr_map_diff_mllm} visualizes the correlation coefficients between every two layers of the LLMs within different MLLMs.
As shown, for all MLLMs, the coefficients are high in almost all regions except the first row and the first column.
The result further enhance our finding about the relative importance of vision tokens.

\subsection{Limitations and future work} \label{subsec:limit_future_work}
We discuss several limitations of our method and provide potential solutions.
First, a general issue of prompt-aware vision token reduction methods is that they require attention scores in the prefilling stage. Thus, they cannot use existing I/O aware approaches like FlashAttention2 for further acceleration.
Note that we can still use those approaches with our method in the training and the decoding stage.
Although Sec.~\ref{sbusec:comp_flashattn} shows that our method can outperform FlashAttention2, we will explore an I/O aware version of our method in the future.
For example, we may get the calculated distances of queries and keys from static random-access memory (SRAM) to replace attention scores in our method.
Second, our method finds optimal reduction strategies on image understanding data, which are not optimal for video understanding.
This is probably because videos are more redundant than images in terms of vision tokens.
A possible solution is to search reduction strategies on video data.
Third, our method decides which token to remove only in the prefilling stage. It is highly possible that as the generation of the response, some vision tokens are no longer essential and can be removed. A potential solution is to adjust the reduction strategy based on the sequential output tokens, as well as the instruction tokens.

%% file: main.bbl
\begin{thebibliography}{55}
\providecommand{\natexlab}[1]{#1}
\providecommand{\url}[1]{\texttt{#1}}
\expandafter\ifx\csname urlstyle\endcsname\relax
  \providecommand{\doi}[1]{doi: #1}\else
  \providecommand{\doi}{doi: \begingroup \urlstyle{rm}\Url}\fi

\bibitem[Bai et~al.(2023)Bai, Bai, Yang, Wang, Tan, Wang, Lin, Zhou, and Zhou]{bai2023qwen}
Jinze Bai, Shuai Bai, Shusheng Yang, Shijie Wang, Sinan Tan, Peng Wang, Junyang Lin, Chang Zhou, and Jingren Zhou.
\newblock Qwen-vl: A versatile vision-language model for understanding, localization, text reading, and beyond.
\newblock 2023.

\bibitem[Biten et~al.(2019)Biten, Tito, Mafla, Gomez, Rusinol, Valveny, Jawahar, and Karatzas]{biten2019scene}
Ali~Furkan Biten, Ruben Tito, Andres Mafla, Lluis Gomez, Mar{\c{c}}al Rusinol, Ernest Valveny, CV Jawahar, and Dimosthenis Karatzas.
\newblock Scene text visual question answering.
\newblock In \emph{Proceedings of the IEEE/CVF international conference on computer vision}, pages 4291--4301, 2019.

\bibitem[Cha et~al.(2024)Cha, Kang, Mun, and Roh]{cha2024honeybee}
Junbum Cha, Wooyoung Kang, Jonghwan Mun, and Byungseok Roh.
\newblock Honeybee: Locality-enhanced projector for multimodal llm.
\newblock In \emph{Proceedings of the IEEE/CVF Conference on Computer Vision and Pattern Recognition}, pages 13817--13827, 2024.

\bibitem[Chefer et~al.(2021)Chefer, Gur, and Wolf]{chefer2021transformer}
Hila Chefer, Shir Gur, and Lior Wolf.
\newblock Transformer interpretability beyond attention visualization.
\newblock In \emph{Proceedings of the IEEE/CVF conference on computer vision and pattern recognition}, pages 782--791, 2021.

\bibitem[Chen et~al.(2024{\natexlab{a}})Chen, Zhao, Liu, Bai, Lin, Zhou, and Chang]{chen2024image}
Liang Chen, Haozhe Zhao, Tianyu Liu, Shuai Bai, Junyang Lin, Chang Zhou, and Baobao Chang.
\newblock An image is worth 1/2 tokens after layer 2: Plug-and-play inference acceleration for large vision-language models.
\newblock In \emph{European Conference on Computer Vision}, 2024{\natexlab{a}}.

\bibitem[Chen et~al.(2024{\natexlab{b}})Chen, Wang, Tian, Ye, Gao, Cui, Tong, Hu, Luo, Ma, et~al.]{chen2024far}
Zhe Chen, Weiyun Wang, Hao Tian, Shenglong Ye, Zhangwei Gao, Erfei Cui, Wenwen Tong, Kongzhi Hu, Jiapeng Luo, Zheng Ma, et~al.
\newblock How far are we to gpt-4v? closing the gap to commercial multimodal models with open-source suites.
\newblock \emph{arXiv preprint arXiv:2404.16821}, 2024{\natexlab{b}}.

\bibitem[Chen et~al.(2024{\natexlab{c}})Chen, Wu, Wang, Su, Chen, Xing, Zhong, Zhang, Zhu, Lu, et~al.]{chen2024internvl}
Zhe Chen, Jiannan Wu, Wenhai Wang, Weijie Su, Guo Chen, Sen Xing, Muyan Zhong, Qinglong Zhang, Xizhou Zhu, Lewei Lu, et~al.
\newblock Internvl: Scaling up vision foundation models and aligning for generic visual-linguistic tasks.
\newblock In \emph{Proceedings of the IEEE/CVF Conference on Computer Vision and Pattern Recognition}, pages 24185--24198, 2024{\natexlab{c}}.

\bibitem[Dai et~al.(2023)Dai, Li, Li, Tiong, Zhao, Wang, Li, Fung, and Hoi]{instructblip}
Wenliang Dai, Junnan Li, Dongxu Li, Anthony Meng~Huat Tiong, Junqi Zhao, Weisheng Wang, Boyang Li, Pascale Fung, and Steven Hoi.
\newblock Instructblip: Towards general-purpose vision-language models with instruction tuning.
\newblock \emph{Advances in neural information processing systems}, 36, 2023.

\bibitem[Dao(2023)]{dao2023flashattention}
Tri Dao.
\newblock {FlashAttention-2}: Faster attention with better parallelism and work partitioning.
\newblock \emph{arXiv preprint arXiv:2307.08691}, 2023.

\bibitem[Dao et~al.(2022)Dao, Fu, Ermon, Rudra, and R{\'e}]{dao2022flashattention}
Tri Dao, Dan Fu, Stefano Ermon, Atri Rudra, and Christopher R{\'e}.
\newblock {FlashAttention}: Fast and memory-efficient exact attention with io-awareness.
\newblock \emph{Advances in Neural Information Processing Systems}, 35:\penalty0 16344--16359, 2022.

\bibitem[Dosovitskiy et~al.(2021)Dosovitskiy, Beyer, Kolesnikov, Weissenborn, Zhai, Unterthiner, Dehghani, Minderer, Heigold, Gelly, Uszkoreit, and Houlsby]{dosovitskiy2021an}
Alexey Dosovitskiy, Lucas Beyer, Alexander Kolesnikov, Dirk Weissenborn, Xiaohua Zhai, Thomas Unterthiner, Mostafa Dehghani, Matthias Minderer, Georg Heigold, Sylvain Gelly, Jakob Uszkoreit, and Neil Houlsby.
\newblock An image is worth 16x16 words: Transformers for image recognition at scale.
\newblock In \emph{International Conference on Learning Representations}, 2021.

\bibitem[Fu et~al.(2024)Fu, Dai, Luo, Li, Ren, Zhang, Wang, Zhou, Shen, Zhang, et~al.]{fu2024video}
Chaoyou Fu, Yuhan Dai, Yondong Luo, Lei Li, Shuhuai Ren, Renrui Zhang, Zihan Wang, Chenyu Zhou, Yunhang Shen, Mengdan Zhang, et~al.
\newblock {Video-MME}: The first-ever comprehensive evaluation benchmark of multi-modal llms in video analysis.
\newblock \emph{arXiv preprint arXiv:2405.21075}, 2024.

\bibitem[Guan et~al.(2024)Guan, Liu, Wu, Xian, Li, Liu, Wang, Chen, Huang, Yacoob, et~al.]{guan2024hallusionbench}
Tianrui Guan, Fuxiao Liu, Xiyang Wu, Ruiqi Xian, Zongxia Li, Xiaoyu Liu, Xijun Wang, Lichang Chen, Furong Huang, Yaser Yacoob, et~al.
\newblock Hallusionbench: an advanced diagnostic suite for entangled language hallucination and visual illusion in large vision-language models.
\newblock In \emph{Proceedings of the IEEE/CVF Conference on Computer Vision and Pattern Recognition}, pages 14375--14385, 2024.

\bibitem[Hudson and Manning(2019)]{hudson2019gqa}
Drew~A Hudson and Christopher~D Manning.
\newblock {GQA}: A new dataset for real-world visual reasoning and compositional question answering.
\newblock In \emph{Proceedings of the IEEE/CVF conference on computer vision and pattern recognition}, pages 6700--6709, 2019.

\bibitem[Jian et~al.(2024)Jian, Gao, and Vosoughi]{jian2024bootstrapping}
Yiren Jian, Chongyang Gao, and Soroush Vosoughi.
\newblock Bootstrapping vision-language learning with decoupled language pre-training.
\newblock \emph{Advances in Neural Information Processing Systems}, 36, 2024.

\bibitem[Kembhavi et~al.(2016)Kembhavi, Salvato, Kolve, Seo, Hajishirzi, and Farhadi]{kembhavi2016diagram}
Aniruddha Kembhavi, Mike Salvato, Eric Kolve, Minjoon Seo, Hannaneh Hajishirzi, and Ali Farhadi.
\newblock A diagram is worth a dozen images.
\newblock In \emph{Computer Vision--ECCV 2016: 14th European Conference, Amsterdam, The Netherlands, October 11--14, 2016, Proceedings, Part IV 14}, pages 235--251. Springer, 2016.

\bibitem[Kendall(1938)]{kendall1938new}
Maurice~G Kendall.
\newblock A new measure of rank correlation.
\newblock \emph{Biometrika}, 30\penalty0 (1-2):\penalty0 81--93, 1938.

\bibitem[Li et~al.(2023{\natexlab{a}})Li, Li, Savarese, and Hoi]{li2023blip}
Junnan Li, Dongxu Li, Silvio Savarese, and Steven Hoi.
\newblock Blip-2: Bootstrapping language-image pre-training with frozen image encoders and large language models.
\newblock In \emph{International conference on machine learning}, pages 19730--19742. PMLR, 2023{\natexlab{a}}.

\bibitem[Li et~al.(2024{\natexlab{a}})Li, Wang, He, Li, Wang, Liu, Wang, Xu, Chen, Luo, et~al.]{li2024mvbench}
Kunchang Li, Yali Wang, Yinan He, Yizhuo Li, Yi Wang, Yi Liu, Zun Wang, Jilan Xu, Guo Chen, Ping Luo, et~al.
\newblock {MVBench}: A comprehensive multi-modal video understanding benchmark.
\newblock In \emph{Proceedings of the IEEE/CVF Conference on Computer Vision and Pattern Recognition}, pages 22195--22206, 2024{\natexlab{a}}.

\bibitem[Li et~al.(2024{\natexlab{b}})Li, Yuan, Liu, Tang, Wang, Zhu, and Zhang]{li2024tokenpacker}
Wentong Li, Yuqian Yuan, Jian Liu, Dongqi Tang, Song Wang, Jianke Zhu, and Lei Zhang.
\newblock Tokenpacker: Efficient visual projector for multimodal llm.
\newblock \emph{arXiv preprint arXiv:2407.02392}, 2024{\natexlab{b}}.

\bibitem[Li et~al.(2023{\natexlab{b}})Li, Du, Zhou, Wang, Zhao, and Wen]{li2023evaluating}
Yifan Li, Yifan Du, Kun Zhou, Jinpeng Wang, Wayne~Xin Zhao, and Ji-Rong Wen.
\newblock Evaluating object hallucination in large vision-language models.
\newblock \emph{arXiv preprint arXiv:2305.10355}, 2023{\natexlab{b}}.

\bibitem[Li et~al.(2024{\natexlab{c}})Li, Huang, Yang, Venkitesh, Locatelli, Ye, Cai, Lewis, and Chen]{li2024snapkv}
Yuhong Li, Yingbing Huang, Bowen Yang, Bharat Venkitesh, Acyr Locatelli, Hanchen Ye, Tianle Cai, Patrick Lewis, and Deming Chen.
\newblock Snapkv: Llm knows what you are looking for before generation.
\newblock \emph{arXiv preprint arXiv:2404.14469}, 2024{\natexlab{c}}.

\bibitem[Li et~al.(2024{\natexlab{d}})Li, Wang, and Jia]{li2024llamavid}
Yanwei Li, Chengyao Wang, and Jiaya Jia.
\newblock {LLaMA-VID}: An image is worth 2 tokens in large language models.
\newblock In \emph{European Conference on Computer Vision}, 2024{\natexlab{d}}.

\bibitem[Li et~al.(2024{\natexlab{e}})Li, Zhang, Wang, Zhong, Chen, Chu, Liu, and Jia]{li2024mini}
Yanwei Li, Yuechen Zhang, Chengyao Wang, Zhisheng Zhong, Yixin Chen, Ruihang Chu, Shaoteng Liu, and Jiaya Jia.
\newblock Mini-gemini: Mining the potential of multi-modality vision language models.
\newblock \emph{arXiv preprint arXiv:2403.18814}, 2024{\natexlab{e}}.

\bibitem[Li et~al.(2024{\natexlab{f}})Li, Yang, Liu, Ma, Zhang, Yang, Sun, Liu, and Bai]{li2024monkey}
Zhang Li, Biao Yang, Qiang Liu, Zhiyin Ma, Shuo Zhang, Jingxu Yang, Yabo Sun, Yuliang Liu, and Xiang Bai.
\newblock Monkey: Image resolution and text label are important things for large multi-modal models.
\newblock In \emph{Proceedings of the IEEE/CVF Conference on Computer Vision and Pattern Recognition}, pages 26763--26773, 2024{\natexlab{f}}.

\bibitem[Lin et~al.(2024)Lin, Lin, Lin, and Ji]{lin2024boosting}
Zhihang Lin, Mingbao Lin, Luxi Lin, and Rongrong Ji.
\newblock Boosting multimodal large language models with visual tokens withdrawal for rapid inference.
\newblock \emph{arXiv preprint arXiv:2405.05803}, 2024.

\bibitem[Liu et~al.(2023)Liu, Li, Wu, and Lee]{liu2023visual}
Haotian Liu, Chunyuan Li, Qingyang Wu, and Yong~Jae Lee.
\newblock Visual instruction tuning.
\newblock \emph{Advances in neural information processing systems}, 36, 2023.

\bibitem[Liu et~al.(2024{\natexlab{a}})Liu, Li, Li, and Lee]{liu2024improved}
Haotian Liu, Chunyuan Li, Yuheng Li, and Yong~Jae Lee.
\newblock Improved baselines with visual instruction tuning.
\newblock In \emph{Proceedings of the IEEE/CVF Conference on Computer Vision and Pattern Recognition}, pages 26296--26306, 2024{\natexlab{a}}.

\bibitem[Liu et~al.(2024{\natexlab{b}})Liu, Li, Li, Li, Zhang, Shen, and Lee]{liu2024llavanext}
Haotian Liu, Chunyuan Li, Yuheng Li, Bo Li, Yuanhan Zhang, Sheng Shen, and Yong~Jae Lee.
\newblock Llava-next: Improved reasoning, ocr, and world knowledge, 2024{\natexlab{b}}.

\bibitem[Liu et~al.(2025)Liu, Duan, Zhang, Li, Zhang, Zhao, Yuan, Wang, He, Liu, et~al.]{liu2025mmbench}
Yuan Liu, Haodong Duan, Yuanhan Zhang, Bo Li, Songyang Zhang, Wangbo Zhao, Yike Yuan, Jiaqi Wang, Conghui He, Ziwei Liu, et~al.
\newblock {MMBench}: Is your multi-modal model an all-around player?
\newblock In \emph{European Conference on Computer Vision}, pages 216--233. Springer, 2025.

\bibitem[Lu et~al.(2023)Lu, Bansal, Xia, Liu, Li, Hajishirzi, Cheng, Chang, Galley, and Gao]{lu2023mathvista}
Pan Lu, Hritik Bansal, Tony Xia, Jiacheng Liu, Chunyuan Li, Hannaneh Hajishirzi, Hao Cheng, Kai-Wei Chang, Michel Galley, and Jianfeng Gao.
\newblock Mathvista: Evaluating mathematical reasoning of foundation models in visual contexts.
\newblock \emph{arXiv preprint arXiv:2310.02255}, 2023.

\bibitem[Luo et~al.(2024)Luo, Zhou, Zhang, Zheng, Sun, and Ji]{luo2024feast}
Gen Luo, Yiyi Zhou, Yuxin Zhang, Xiawu Zheng, Xiaoshuai Sun, and Rongrong Ji.
\newblock Feast your eyes: Mixture-of-resolution adaptation for multimodal large language models.
\newblock \emph{arXiv preprint arXiv:2403.03003}, 2024.

\bibitem[Masry et~al.(2022)Masry, Long, Tan, Joty, and Hoque]{masry-etal-2022-chartqa}
Ahmed Masry, Do Long, Jia~Qing Tan, Shafiq Joty, and Enamul Hoque.
\newblock {C}hart{QA}: A benchmark for question answering about charts with visual and logical reasoning.
\newblock In \emph{Findings of the Association for Computational Linguistics: ACL 2022}, pages 2263--2279, Dublin, Ireland, 2022. Association for Computational Linguistics.

\bibitem[Mathew et~al.(2021)Mathew, Karatzas, and Jawahar]{mathew2021docvqa}
Minesh Mathew, Dimosthenis Karatzas, and CV Jawahar.
\newblock {DocVQA}: A dataset for vqa on document images.
\newblock In \emph{Proceedings of the IEEE/CVF winter conference on applications of computer vision}, pages 2200--2209, 2021.

\bibitem[Mockus(2005)]{mockus2005bayesian}
Jonas Mockus.
\newblock The bayesian approach to global optimization.
\newblock In \emph{System Modeling and Optimization: Proceedings of the 10th IFIP Conference New York City, USA, August 31--September 4, 1981}, pages 473--481. Springer, 2005.

\bibitem[Radford et~al.(2021)Radford, Kim, Hallacy, Ramesh, Goh, Agarwal, Sastry, Askell, Mishkin, Clark, et~al.]{radford2021learning}
Alec Radford, Jong~Wook Kim, Chris Hallacy, Aditya Ramesh, Gabriel Goh, Sandhini Agarwal, Girish Sastry, Amanda Askell, Pamela Mishkin, Jack Clark, et~al.
\newblock Learning transferable visual models from natural language supervision.
\newblock In \emph{International conference on machine learning}, pages 8748--8763. PMLR, 2021.

\bibitem[Shang et~al.(2024)Shang, Cai, Xu, Lee, and Yan]{shang2024llava}
Yuzhang Shang, Mu Cai, Bingxin Xu, Yong~Jae Lee, and Yan Yan.
\newblock Llava-prumerge: Adaptive token reduction for efficient large multimodal models.
\newblock \emph{arXiv preprint arXiv:2403.15388}, 2024.

\bibitem[Singh et~al.(2019)Singh, Natarajan, Shah, Jiang, Chen, Batra, Parikh, and Rohrbach]{singh2019towards}
Amanpreet Singh, Vivek Natarajan, Meet Shah, Yu Jiang, Xinlei Chen, Dhruv Batra, Devi Parikh, and Marcus Rohrbach.
\newblock Towards {VQA} models that can read.
\newblock In \emph{Proceedings of the IEEE/CVF conference on computer vision and pattern recognition}, pages 8317--8326, 2019.

\bibitem[Tong et~al.(2024)Tong, Brown, Wu, Woo, Middepogu, Akula, Yang, Yang, Iyer, Pan, et~al.]{tong2024cambrian}
Shengbang Tong, Ellis Brown, Penghao Wu, Sanghyun Woo, Manoj Middepogu, Sai~Charitha Akula, Jihan Yang, Shusheng Yang, Adithya Iyer, Xichen Pan, et~al.
\newblock Cambrian-1: A fully open, vision-centric exploration of multimodal llms.
\newblock \emph{arXiv preprint arXiv:2406.16860}, 2024.

\bibitem[xAI(2024)]{RealWordQA}
xAI.
\newblock {RealWordQA}: \url{https://huggingface.co/datasets/xai-org/RealworldQA}, 2024.

\bibitem[Xiao et~al.(2024)Xiao, Tian, Chen, Han, and Lewis]{xiao2024efficient}
Guangxuan Xiao, Yuandong Tian, Beidi Chen, Song Han, and Mike Lewis.
\newblock Efficient streaming language models with attention sinks.
\newblock In \emph{The Twelfth International Conference on Learning Representations}, 2024.

\bibitem[xiaoju ye(2023)]{calflops}
xiaoju ye.
\newblock {calflops}: a flops and params calculate tool for neural networks in pytorch framework, 2023.

\bibitem[Xing et~al.(2024)Xing, Huang, Dong, Lu, Zhang, Zang, Cao, He, Wang, Wu, et~al.]{xing2024pyramiddrop}
Long Xing, Qidong Huang, Xiaoyi Dong, Jiajie Lu, Pan Zhang, Yuhang Zang, Yuhang Cao, Conghui He, Jiaqi Wang, Feng Wu, et~al.
\newblock Pyramiddrop: Accelerating your large vision-language models via pyramid visual redundancy reduction.
\newblock \emph{arXiv preprint arXiv:2410.17247}, 2024.

\bibitem[Xu et~al.(2024)Xu, Yao, Guo, Cui, Ni, Ge, Chua, Liu, Sun, and Huang]{xu2024llava}
Ruyi Xu, Yuan Yao, Zonghao Guo, Junbo Cui, Zanlin Ni, Chunjiang Ge, Tat-Seng Chua, Zhiyuan Liu, Maosong Sun, and Gao Huang.
\newblock Llava-uhd: an lmm perceiving any aspect ratio and high-resolution images.
\newblock \emph{arXiv preprint arXiv:2403.11703}, 2024.

\bibitem[Yao et~al.(2024)Yao, Li, Ren, Wang, Liu, Sun, and Hou]{yao2024deco}
Linli Yao, Lei Li, Shuhuai Ren, Lean Wang, Yuanxin Liu, Xu Sun, and Lu Hou.
\newblock Deco: Decoupling token compression from semantic abstraction in multimodal large language models.
\newblock \emph{arXiv preprint arXiv:2405.20985}, 2024.

\bibitem[Ye et~al.(2024)Ye, Gan, Huang, Ge, Shan, and Tang]{ye2024voco}
Xubing Ye, Yukang Gan, Xiaoke Huang, Yixiao Ge, Ying Shan, and Yansong Tang.
\newblock {VoCo-LLaMA}: Towards vision compression with large language models.
\newblock \emph{arXiv preprint arXiv:2406.12275}, 2024.

\bibitem[Yue et~al.(2024)Yue, Ni, Zhang, Zheng, Liu, Zhang, Stevens, Jiang, Ren, Sun, et~al.]{yue2024mmmu}
Xiang Yue, Yuansheng Ni, Kai Zhang, Tianyu Zheng, Ruoqi Liu, Ge Zhang, Samuel Stevens, Dongfu Jiang, Weiming Ren, Yuxuan Sun, et~al.
\newblock Mmmu: A massive multi-discipline multimodal understanding and reasoning benchmark for expert agi.
\newblock In \emph{Proceedings of the IEEE/CVF Conference on Computer Vision and Pattern Recognition}, pages 9556--9567, 2024.

\bibitem[Zhang et~al.(2024{\natexlab{a}})Zhang, Li, Zhang, Pu, Cahyono, Hu, Liu, Zhang, Yang, Li, and Liu]{zhang2024lmmsevalrealitycheckevaluation}
Kaichen Zhang, Bo Li, Peiyuan Zhang, Fanyi Pu, Joshua~Adrian Cahyono, Kairui Hu, Shuai Liu, Yuanhan Zhang, Jingkang Yang, Chunyuan Li, and Ziwei Liu.
\newblock {LMMs-Eval}: Reality check on the evaluation of large multimodal models, 2024{\natexlab{a}}.

\bibitem[Zhang et~al.(2011)Zhang, Zhang, Mou, and Zhang]{zhang2011fsim}
Lin Zhang, Lei Zhang, Xuanqin Mou, and David Zhang.
\newblock Fsim: A feature similarity index for image quality assessment.
\newblock \emph{IEEE transactions on Image Processing}, 20\penalty0 (8):\penalty0 2378--2386, 2011.

\bibitem[Zhang et~al.(2024{\natexlab{b}})Zhang, Wen, Fu, Wang, Zhang, Wang, and Jin]{zhang2024beyond}
Yi-Fan Zhang, Qingsong Wen, Chaoyou Fu, Xue Wang, Zhang Zhang, Liang Wang, and Rong Jin.
\newblock Beyond llava-hd: Diving into high-resolution large multimodal models.
\newblock \emph{arXiv preprint arXiv:2406.08487}, 2024{\natexlab{b}}.

\bibitem[Zhao et~al.(2020)Zhao, Zhang, Huang, Shen, and Zhao]{zhao2020dehazing}
Shiyu Zhao, Lin Zhang, Shuaiyi Huang, Ying Shen, and Shengjie Zhao.
\newblock Dehazing evaluation: Real-world benchmark datasets, criteria, and baselines.
\newblock \emph{IEEE Transactions on Image Processing}, 29:\penalty0 6947--6962, 2020.

\bibitem[Zhao et~al.(2022)Zhao, Zhang, Schulter, Zhao, Vijay~Kumar, Stathopoulos, Chandraker, and Metaxas]{zhao2022exploiting}
Shiyu Zhao, Zhixing Zhang, Samuel Schulter, Long Zhao, BG Vijay~Kumar, Anastasis Stathopoulos, Manmohan Chandraker, and Dimitris~N Metaxas.
\newblock Exploiting unlabeled data with vision and language models for object detection.
\newblock In \emph{European conference on computer vision}, pages 159--175. Springer, 2022.

\bibitem[Zhao et~al.(2024{\natexlab{a}})Zhao, Schulter, Zhao, Zhang, Suh, Chandraker, and Metaxas]{zhao2024taming}
Shiyu Zhao, Samuel Schulter, Long Zhao, Zhixing Zhang, Yumin Suh, Manmohan Chandraker, and Dimitris~N Metaxas.
\newblock Taming self-training for open-vocabulary object detection.
\newblock In \emph{Proceedings of the IEEE/CVF Conference on Computer Vision and Pattern Recognition}, pages 13938--13947, 2024{\natexlab{a}}.

\bibitem[Zhao et~al.(2024{\natexlab{b}})Zhao, Zhao, Suh, Metaxas, Chandraker, and Schulter]{zhao2024generating}
Shiyu Zhao, Long Zhao, Yumin Suh, Dimitris~N Metaxas, Manmohan Chandraker, and Samuel Schulter.
\newblock Generating enhanced negatives for training language-based object detectors.
\newblock In \emph{Proceedings of the IEEE/CVF Conference on Computer Vision and Pattern Recognition}, pages 13592--13602, 2024{\natexlab{b}}.

\bibitem[Zhu et~al.(2024)Zhu, Chen, Shen, Li, and Elhoseiny]{zhu2023minigpt}
Deyao Zhu, Jun Chen, Xiaoqian Shen, Xiang Li, and Mohamed Elhoseiny.
\newblock Mini{GPT}-4: Enhancing vision-language understanding with advanced large language models.
\newblock In \emph{The Twelfth International Conference on Learning Representations}, 2024.

\end{thebibliography}
